\pdfoutput=1
\documentclass[11pt]{article}

\usepackage{EMNLP2022}

\usepackage{times}
\usepackage{latexsym}

\usepackage[T1]{fontenc}
\usepackage[utf8]{inputenc}

\usepackage{microtype}

\usepackage{inconsolata}

\usepackage{booktabs}
\usepackage{graphicx}
\usepackage{subcaption}
\usepackage{tcolorbox}
\usepackage{makecell}
\usepackage{tabularx}
\usepackage{enumitem}
\usepackage{hyperref}
\usepackage[noabbrev,capitalize,nameinlink]{cleveref}
\usepackage{tikz}
\usepackage{tikz-dependency}
\usepackage{adjustbox}
\usepackage{todonotes}
\usepackage[utf8]{inputenc}
\usepackage{fontawesome5}

\newcommand*\circled[1]{\tikz[baseline=(char.base)]{
            \node[shape=circle,draw,inner sep=.6pt] (char) {#1};}}
            
\usepackage{url}

\usepackage{breakurl}

\title{Experimental Standards for Deep Learning in\\ Natural Language Processing Research}

\author{Dennis Ulmer\textsuperscript{\faCompass} \hspace{.5em}  Elisa Bassignana\textsuperscript{\faCompass} \hspace{.5em}  Max M\"uller-Eberstein\textsuperscript{\faCompass} \hspace{.5em}  Daniel Varab\textsuperscript{\faCompass} \\  {\bf Mike Zhang}\textsuperscript{\faCompass} \hspace{.5em} {\bf Rob van der Goot}\textsuperscript{\faCompass} \hspace{.5em} {\bf Christian Hardmeier}\textsuperscript{\faCompass} \hspace{.5em}  {\bf Barbara Plank}\textsuperscript{\faCompass}\textsuperscript{\faMountain}\textsuperscript{\faRobot} \\
        \textsuperscript{\faCompass}Department of Computer Science, IT University of Copenhagen, Denmark \\ 
        \textsuperscript{\faMountain}Center for Information and Language Processing (CIS), LMU Munich, Germany \\
        \textsuperscript{\faRobot}Munich Center for Machine Learning (MCML), Munich, Germany \\
        \texttt{dennis.ulmer@mailbox.org}}

\begin{document}
\maketitle
\begin{abstract}
\looseness=-1
     The field of Deep Learning (DL) has undergone explosive growth during the last decade, with a substantial impact on Natural Language Processing (NLP) as well.
     Yet, compared to more established disciplines, a lack of common experimental standards remains an open challenge to the field at large. % I dont like the flow of the line below to be honest
     %Yet, a lack of common experimental standards compared to more established disciplines has remained an open challenge to the field at large.
     % Yet, as with other fields employing DL techniques, there has been a lack of common experimental standards compared to more established disciplines. 
     Starting from fundamental scientific principles, we distill ongoing discussions on experimental standards in NLP into a single, widely-applicable methodology. Following these best practices is crucial to strengthen experimental evidence, improve reproducibility and support scientific progress. These standards are further collected in a public repository to help them transparently adapt to future needs.
\end{abstract}

%EMNLP special theme: Open questions, major obstacles, and unresolved issues in NLP”
\section{Introduction}
% Aim: Collection of best practice from all the papers we read, to become a sort of guide for beginning PhD students + tips for seasoned researchers alike

\begin{figure}[t]
    \centering 
    \includegraphics[width=.995\linewidth]{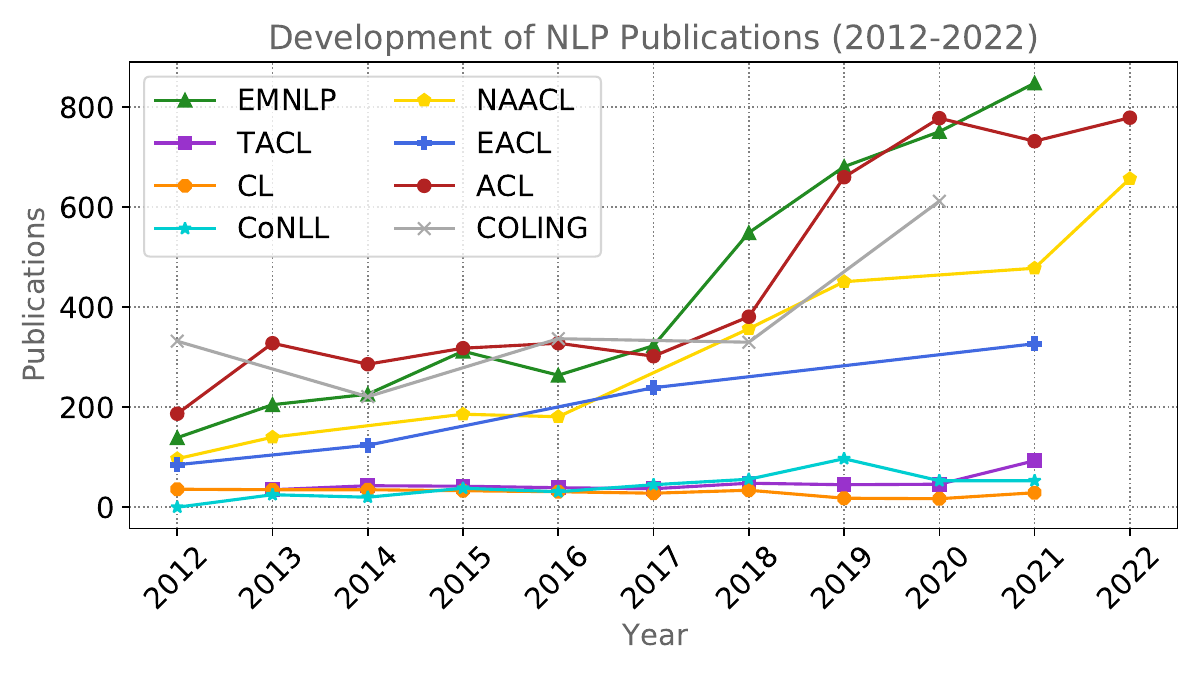}
    \looseness=-1
    \caption{\textbf{Development of NLP publications.} Shown is the development of NLP measured by the number of peer-reviewed publications between 2012--2022 based on the data collected by~\citet{rei2022publications}.}
    \label{fig:development-nlp}
\end{figure}

Spurred by the advances in Machine Learning (ML) and Deep Learning (DL), the field of Natural Language Processing (NLP) has seen immense growth over the span of the last ten years, as illustrated by the number of publications in \cref{fig:development-nlp}. 
%\cref{fig:development-ai-dl} shows how, according to \citet{zhang2021ai}, the number of peer-reviewed papers has increased twelve-fold compared to the year 2000. At the same time, interest in Deep Learning (DL) has increased substantially as well, demonstrated via Google Trends in the same figure. 
While such progress is remarkable, rapid growth comes at a cost: Akin to concerns in other disciplines \citep{john2012measuring, jensen2021there}, several authors have noted major obstacles with reproducibility \citep{gundersen2018state, belz2021systematic} and a lack of significance testing \citep{marie2021scientific} or published results not carrying over to different experimental setups, for instance in text generation \citep{gehrmann2022repairing} and with respect to new model architectures \citep{narang2021transformer}.
%Reinforcement Learning \citep{henderson2018deep, agarwal2021deep}, and optimization \citep{schmidt2021descending}.
Others have questioned commonly-accepted procedures \citep{gorman2019we, sogaard2021we, bouthillier2021accounting, groot2021we} as well as the (negative) impacts of research on society \citep{hovy-spruit-2016-social, mohamed2020decolonial, bender2021dangers, birhane2021values} and environment \citep{strubell-etal-2019-energy, schwartz2020green, henderson2020towards}. These problems have not gone unnoticed---many of the mentioned works have proposed a cornucopia of solutions. In a quickly-moving environment however, keeping track and implementing these proposals becomes challenging. In this work, we weave these open issues together into a cohesive methodology for gathering stronger experimental evidence, that can be implemented with reasonable effort.

\begin{figure}
    \centering
    \includegraphics[width=\columnwidth]{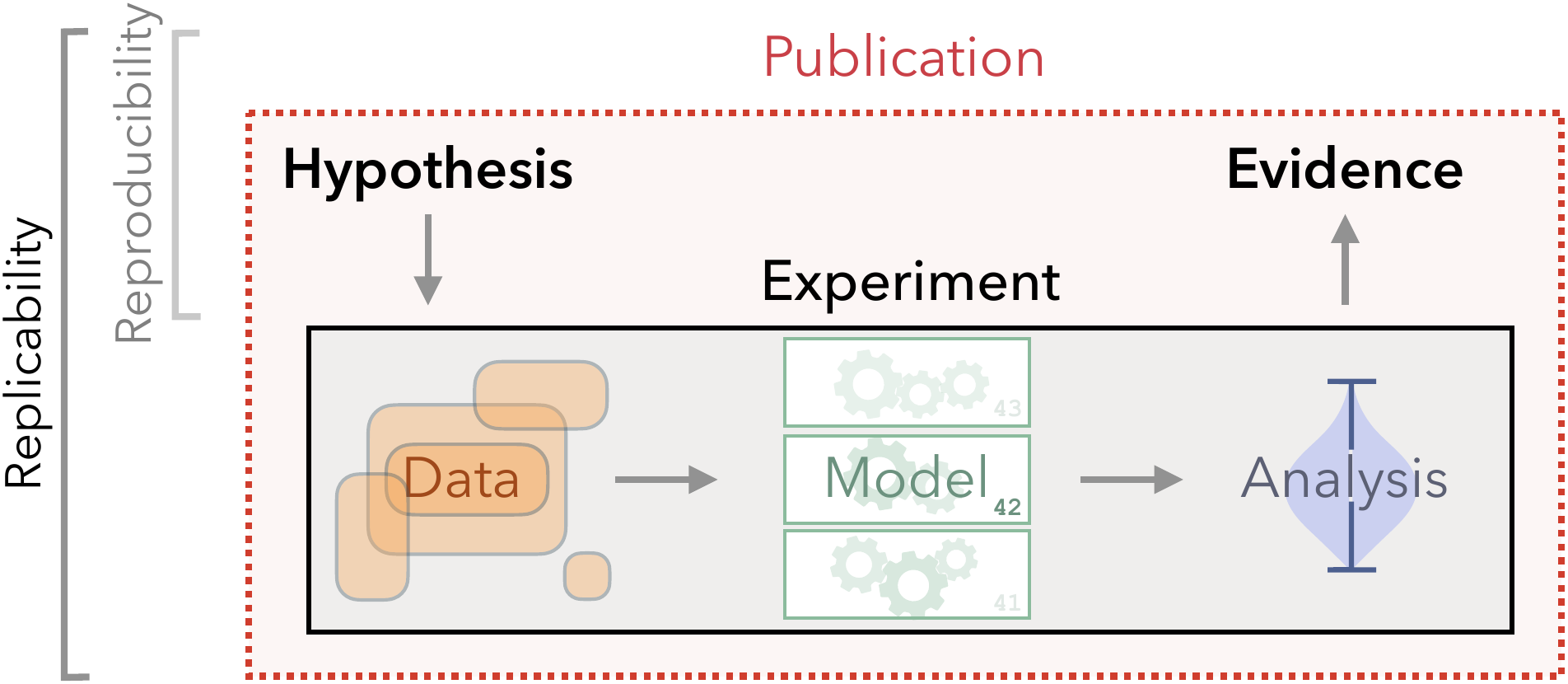}
    \looseness=-1
    \caption{\textbf{Visualization of the Scientific Process in Deep Learning.} Uncertainty is introduced at each step, influencing the resulting evidence as well as the documentation required for reproducibility or replicability.}
    \label{fig:process}
\end{figure}

Based on the scientific method (\cref{sec:preliminaries}), we divide the empirical research process---obtaining evidence from data via modeling---into four steps, which are depicted in \cref{fig:process}: \emph{Data} (\cref{sec:data}), including dataset creation and usage, \emph{Codebase \& Models} (\cref{sec:model-codebase}), \emph{Experiments \& Analysis} (\cref{sec:experiments-analysis}) and \emph{Publication} (\cref{sec:publication}). For each step, we survey contemporary findings and summarize them into actionable practices for empirical research. 
% While written mostly from the perspective of NLP researchers, we expect many of these insights will be useful for practitioners of other adjacent sub-fields of ML and DL.
Using insights from adjacent sub-fields of ML / DL, we extract useful insights to help overcome current challenges with replicability in NLP.

% \begin{figure*}[t]
%     \centering
%     % \begin{subfigure}[t]{0.48\textwidth}
%         % \centering
%         \includegraphics[width=.48\linewidth]{figures/publications.pdf}
%         \hspace{0.5em}
%         \includegraphics[width=.48\linewidth]{figures/trends.pdf}
%         \looseness=-1
%         % \caption{Development of peer-review AI publications between 2000 and 2019, compiled by \citet{zhang2021ai}.}\label{fig:ai-development}
%     % \end{subfigure}
%     % \hfill
%     % \begin{subfigure}[t]{0.48\textwidth}
%         % \centering
%         % \includegraphics[width=\columnwidth]{./img/trends.pdf}
%         % \looseness=-1
%         % \caption{Development of worldwide interest in the topic \emph{Deep Learning} by Google Trends from 2004 to 2021.}\label{fig:ai-google-trends}
%     % \end{subfigure}
%     \caption{\textbf{Development of AI and DL.} Shown is the development of AI and DL measured by the number of peer-reviewed publications between 2000-2019~\cite{zhang2021ai} on the left and relative interest measured by the search engine Google between 2004-2021~\cite{gtrend200421} on the right. Both experience an explosive growth around 2014.}
%     \label{fig:development-ai-dl}
% \end{figure*}

\paragraph{Contributions}
\circled{1} We survey and summarize a wide array of proposals regarding the improvement of the experimental (and publishing) pipeline in NLP research into a single accessible methodology applicable for a wide and diverse readership. At the end of every section, we provide a summary with the most important points, marked with $\diamond$ to indicate that they should be seen as a minimal requirement, and $\star$ for additional recommended actions.
\circled{2} We create, point to, or supply useful resources to support everyday research activities and improve soundness of research in the field. We furthermore provide examples and case studies illustrating these methods in \cref{app:case-studies}. We also provide an additional list of resources in \cref{app:resources}. The same collection as well as checklists derived from the actionable points at the end of sections are also maintained in an open-source repository,\footnote{\url{https://github.com/Kaleidophon/experimental-standards-deep-learning-research}} 
and we invite the research community to discuss, modify and extend these resources.
\circled{3} We discuss current trends and their implications, hoping to initiate a more widespread conversation about them in the NLP community to facilitate common standards and improve the quality of research.

% \begin{itemize}
%     \itemsep0em
%     \item We survey and summarize a wide array of proposals regarding the improvement of the experimental (and publishing) pipeline into an accessible format in order to make them available to a wide and diverse readership.
 %   \item We create, point to, or supply useful resources to support everyday research activities and improve replicability in the field. At the end of every section, we also provide a summary with the most important points, marked with $\diamond$ to indicate that they should be seen as a minimal requirement, and $\star$ for recommended actions.
%     \item We discuss current trends and implications, hoping to initiate a more widespread discussion about the topic in the ML community to facilitate common standards and improve the quality of research in general.
% \end{itemize}

\section{Preliminaries}\label{sec:preliminaries}

Our proposed methodology must be built on the scientific principles for generating strong evidence for the general advancement of knowledge, as defined by the following terms:

%The scientific method is generally defined as a framework to iteratively expand human knowledge as well as the accuracy thereof \citep{popper1934}. Knowledge in turn is built upon hypotheses supported by strong evidence. These hypotheses must be falsifiable, meaning that \emph{anyone} must be able to evaluate them in light of new evidence. 

\paragraph{The Scientific Method} Knowledge can be obtained through several ways including theory building, qualitative methods, and empirical research \citep{kuhn1970structure, simon1995artificial}. Here, we focus on the latter aspect, in which (exploratory) analyses lead to falsifiable hypotheses that can be tested and iterated upon \citep{popper1934}.\footnote{While such hypothesis-driven science is not always applicable or possible \citep{carroll2019beyond}, it is a strong common denominator that encompasses most empirical ML research.} This process requires that \emph{anyone} must be able to back or dispute these  hypotheses in the light of new evidence.

In the following, we focus on the evidence-based evaluation of hypotheses and how to ensure the scientific soundness of the experiments which gave rise to the original empirical evidence, with a focus on \emph{replicability} and \emph{reproducibility}. In computational literature, one term requires access to the original code and data in order to re-run experiments exactly, while the other requires sufficient information in order to reproduce the original findings even in the absence of code and original data (see also Figure~\ref{fig:process}).\footnote{Strikingly, these central terms already lack agreed-upon definitions \citep{peng2011reproducible,fokkens-etal-2013-offspring,liberman2015replicability,cohen2018three}, however we follow the prevailing definitions in the NLP community~\citep{drummond2009repl, emnlp_rep} as the underlying ideas are equivalent.}

\paragraph{Replicability} Within DL, we take replicability to mean the exact replication of prior reported evidence. In a computational environment, access to the same data, code and tooling should be sufficient to generate prior results. However, many factors, such as hardware differences, make exact replication difficult to achieve.
Nonetheless, we regard experiments to be replicable if a practitioner is able to re-run them to produce the same evidence within a small margin of error dependent on the environment, without the need to approximate or guess experimental details.

\paragraph{Reproducibility} In comparison, we take reproducibility to mean the availability of all necessary and sufficient information such that an experiment's findings can be independently reaffirmed when the same research question is asked. As discussed later, the availability of all components for replicability is rare---even in a computational setting. An experiment then is reproducible if anyone with access to the publication is able to re-identify the original evidence, i.e. exact results differing, but patterns across experiments being equivalent.

\paragraph{} We assume that the practitioner aims to follow these principles in order to find answers to a well-motivated research question by gathering the strongest possible evidence for or against their hypotheses.
The following methods therefore aim to reduce uncertainty in each step of the experimental pipeline in order to ensure reproducibility and/or replicability (visualized in \cref{fig:process}).

\section{Data}\label{sec:data}

Frequently, it is claimed that a model solves a particular cognitive task, however in reality it merely scores higher than others on some specific dataset according to some predefined metric \citep{schlangen-2021-targeting}. Of course, the broader goal is to improve systems more generally by using individual datasets as proxies. Admitting that our experiments cover only a small slice of the real-world sample space will help more transparently measure progress towards this goal. In light of these limitations and as there will always be private or otherwise unavailable datasets which violate replicability, a practitioner must ask themselves: \emph{Which key information about the data must be known in order to reproduce an experiment's findings?} In this section we define requirements for putting this question into practice during dataset creation and usage such that anyone can draw the appropriate conclusions from a published experiment.

\paragraph{Choice of Dataset} The choice of dataset will arise from the need to answer a specific research question within the limits of the available resources. Such answers typically come in the form of comparisons between different experimental setups while using the equivalent data and evaluation metrics.
Using a publicly available, well-documented dataset will likely yield more comparable work, and thus stronger evidence. In absence of public data, creating a new dataset according to guidelines which closely follow prior work can also allow for useful comparisons. Should the research question be entirely unexplored, creating a new dataset will be necessary.
In any case, the data itself must contain the information necessary to generate evidence for the researcher's hypothesis. For example, a model for a classification task will not be learnable unless there are distinguishing characteristics between data points and consistent labels for evaluation. Therefore, an exploratory data analysis is recommended for assessing data quality and anticipating problems with the research setup. Simple baseline methods such as regression analyses or simply manually verifying random samples of the data may provide indications regarding the suitability and difficulty of the task and associated dataset~\citep{caswell2021quality}.
%The importance of EDA was recently highlighted by \citet{caswell2021quality} who found critical mistakes in multilingual corpora --- to the point that for over 40\% of the languages, more than 50\% of the data was judged erroneous (e.g., not representing the reported language). It is thus recommended to have a look at the data before using or releasing it.

\paragraph{Metadata} At a higher level, data sheets and statements \citep{gebru2020,bender2018data} aim to standardize metadata for dataset authorship in order to inform future users about assumptions and potential biases during all levels of data collection and annotation---including the research design~\cite{hovyshrimai21}. Simultaneously, they encourage reflection on whether the authors are adhering to their own guidelines \citep{waseem2021disembodied}. 
%Standard metadata for any given dataset includes its sources and number of unique instances. Should the source be prone to changes over time, a data collection time frame may also be reported.
Generally, higher-level documentation should aim to capture the dataset's \emph{representativeness} with respect to the global population. This is especially crucial for ``high-stakes'' environments in which subpopulations may be disadvantaged due to biases during data collection and annotation \citep{he2019practical, sap2021annotators}. Even in lower--stake scenarios, a model trained on only a subset of the global data distribution can have inconsistent behaviour when applied to a different target data distribution \citep{amour2011underspecification,wilds2020}. For instance, domain differences have a noticeable impact on model performance \citep{white2021examining,rameshkashyap2021}. Increased data diversity can improve the ability of models to generalize to new domains and languages \citep{benjamin2018}, however diversity is difficult to quantify \citep{gong2019} and full coverage is unachievable. This highlights the importance of documenting representativeness in order to ensure reproducibility---even in absence of the original data.
For replicability using the original data, further considerations include long-term storage and versioning, as to ensure equal comparisons in future work (see \cref{app:case-studies-data} for case studies).

\paragraph{Instance Annotation} 
Achieving high data quality entails that the data must be accurate and relevant for the task to enable effective learning~\citep{pustejovsky2012natural,bestpractices} and reliable evaluation~\cite{bowman-dahl-2021-will,basile2021we}. Since most datasets involve human annotation, a careful annotation design is crucial~\cite{pustejovsky2012natural,paun2022statistical}. Ambiguity in natural language poses inherent challenges and disagreement is genuine ~\cite{basile2021we,lucia-keynote,uma2021learning}. As insights into the annotation process are valuable, yet often inaccessible, we recommend to release datasets with individual-coder annotations, as also put forward by~\newcite{basile2021we,prabhakaran-etal-2021-releasing} and to complement data with insights like statistics on inter-annotator coding~\cite{paun2022statistical}, e.g., over time~\citep{braggaar-van-der-goot-2021-challenges}, or coder uncertainty~\cite{elisaspaper}. When creating new datasets such information strengthens the reproducibility of future findings, as they transparently communicate the inherent variability instead of obscuring it.
%In fact, an emerging line of work in NLP and Computer Vision aims at devising methods that treat disagreement as a learning signal, rather than noise~\cite{plank-et-al:EACL14:learning,aroyo2015truth,pavlick-kwiatkowski-2019-inherent,leonardelli-etal-2021-agreeing,popovic-2021-agree}. For a survey, we refer to~\citet{uma2021learning}.

\paragraph{Pre-processing} Given a well-constructed or well-chosen dataset, the first step of an experimental setup will be the process by which a model takes in the data. This must be well documented or replicated---most easily by publishing the associated code---as perceivably tiny pre-processing choices can lead to huge accuracy discrepancies~\citep{pedersen-2008-last,fokkens-etal-2013-offspring}. Typically, this involves decisions such as sentence segmentation, tokenization and normalization. In general, the data setup pipeline should ensure that a model ``observes'' the same kind of data across comparisons.
Next, the dataset must be split into representative subsamples which should only be used for their intended purpose, i.e., model training, tuning and evaluation (see \cref{sec:experiments-analysis}). In order to support claims about the generality of the results, it is necessary to use a test split without overlap with other splits. Alternatively, a tuning / test set could consist of data that is completely foreign to the original dataset \citep{ye2021ood}, ideally even multiple sets~\cite{bouthillier2021accounting}. It should be noted that even separate, static test splits are prone to unconscious ``overfitting'' if they have been in use for a longer period of time, as people aim to beat a particular benchmark \citep{gorman2019we}. If a large variety of resources are not available, it is also possible to construct challenging test sets from existing data \citep{ribeiro-etal-2020-beyond,kiela-etal-2021-dynabench,sogaard2021we}.
Finally, the metrics by which models are evaluated should be consistent across experiments and thus benefit from standardized evaluation code \citep{dehghani2021benchmark}. For some tasks, metrics may be driven by community standards and are well-defined (e.g., classification accuracy). In other cases, approximations must stand in for human judgment (e.g., in machine translation). In either case---but especially in the latter---dataset authors should inform users about desirable performance characteristics and recommended metrics.

\paragraph{Appropriate Conclusions} The results a model achieves on a given data setup should first and foremost be taken as just that. Appropriate, broader conclusions can be drawn using this evidence provided that biases or incompleteness of the data are addressed (e.g., results only being applicable to a subpopulation).
Even with statistical tests for the significance of comparisons, properties such as the size of the dataset and the distributional characteristics of the evaluation metric may influence the statistical power of any evidence gained from experiments~\citep{card2020with}. It is therefore important to keep in mind that in order to claim the reliability of the obtained evidence, for example, larger performance differences are necessary on less data than what might suffice for a large dataset, or across multiple comparisons (see~\cref{sec:experiments-analysis}).
Finally, a practitioner should be aware that a model's ability to achieve high scores on a certain dataset may not be directly attributable to its capability of simulating a cognitive ability, but rather due to spurious correlations in the input~\citep{ilyas2019adversarial, schlangen-2021-targeting, nagarajan2021understanding}. By for instance only exposing models to a subset of features that should be inadequate to solve the task, we can sometimes detect when they take unexpected shortcuts~\citep{fokkens-etal-2013-offspring,zhou2015simple}. Communicating the limits of the data helps future work in reproducing prior findings more accurately.

\definecolor{betterorange}{HTML}{EBAA65}
\begin{tcolorbox}[
    title=\centering{\textcolor{black}{Best Practices: \textbf{Data}}}, 
    colback=white,
    colframe=betterorange, 
    left=2pt, 
    right=8pt, 
    enlarge top by=5pt,
    boxsep=2pt,
    coltext=black
]

  \begin{itemize}[leftmargin=12pt]
      \small
      \itemsep0em
      \item[$\diamond$] Consider dataset and experiment limitations when drawing conclusions \citep{schlangen-2021-targeting};
      \item[$\diamond$] Document task adequacy, representativeness and pre-processing \citep{bender2018data};
      \item[$\diamond$] Split the data such as to avoid spurious correlations;
      \item[$\diamond$] Publish the dataset accessibly \& indicate changes;
\item[$\star$] Perform exploratory data analyses to ensure task adequacy \citep{caswell2021quality};
       \item[$\star$] Publish the dataset with individual-coder annotations, besides aggregation;
      \item[$\star$] Claim significance considering the dataset's statistical power \citep{card2020with}.
  \end{itemize}
\end{tcolorbox}

\section{Codebase \& Models}\label{sec:model-codebase}

\looseness=-1
The NLP community has historically taken pride in promoting open access to papers, data, code, and documentation, but some have also noted room for improvement~\cite{wieling2018reproducibility,belz-etal-2020-reprogen}. One practice has been to open-source all components of the experimental procedure in a repository, consisting of all code, necessary scripts, and detailed documentation. The benefit of such a repository is in its ability to enable direct \emph{replication}. In particular, a comprehensive code base directly enables replicability.
%In contrast, an incomplete code base pushes research towards reproducibility. 
In practice, such documentation is often communicated through a \texttt{README} file, in which user-oriented information is described.\footnote{In~\cref{app:readme}, we propose minimal requirements for a \texttt{README} file and give pointers on files and code structure.} In DL, full datasets can be large and impractical to share. Due to their importance however, it is essential to carefully consider how one can share the data with researchers in the future. Therefore, repositories for long-term data storage backed by public institutions should be preferred (e.g., LINDAT/CLARIN by~\citealp{varadi2008clarin}, more examples in~\cref{app:resources}). Nevertheless, practitioners often can not distribute data due to privacy, legal, or storage reasons. In such cases, practitioners must instead carefully consider how to distribute data and tools to allow future research to produce accurate replications of the original data~\cite{zong2020}. 

%Next to the README, we expect to have access to the data the experiments have been conducted on. Datasets are typically large and cannot be hosted in a code repository. As an example, GitHub strongly recommends a repository that is smaller than 5GB and ideally less than 1GB. Files on GitHub are limited to 100MB each.\footnote{As of 23 Nov., 2021} For reference, the GLUE benchmark~\cite{wang2019glue} takes up 1.2GB. Public repositories for long-term data storage will likely always be available and can be used to host or retrieve data in an accurate manner. Such example could be LINDAT/CLARIN~\cite{varadi2008clarin},\footnote{\url{https://lindat.cz/}} European Language Resources Association (ELRA),\footnote{\url{http://www.elra.info/en/}} Zenodo,\footnote{\url{https://zenodo.org/}} or more recently: \texttt{HuggingFace} \texttt{datasets}~\cite{quentin_lhoest_2021_5510481}.
%Once the project or paper is submitted for review, a code base can be attached in terms of a compressed archive (e.g., \texttt{.zip}) file or a anonymized code repository link. To prevent desk rejects due to conflicting anonymity, a recent practice is to anonymize their repository (example in~\cref{app:resources}).

% \todo{elaborate on how we can't always share everything and explain what to do then.. perhaps even explain zong et al.}

\paragraph{Hyperparameter Search} Hyperparameter tuning strategies remain an open area of research (e.g., \citealp{bischl2021hyperparameter}), but are central to the replication of contemporary models. The following rules of thumb exist: Grid search or Bayesian optimization can be applied if few parameters can be searched exhaustively under the computation budget. Otherwise, random search is preferred, as it explores the search space more efficiently~\citep{bergstra2012random}. Advanced methods like Bayesian Optimization~\citep{snoek2012practical} and bandit search-based approaches~\citep{li2017hyperband} can be used as well if applicable~\citep{bischl2021hyperparameter}. To avoid unnecessary guesswork, the following information is expected: Hyperparameters that were searched per model (including options and ranges), the final hyperparameter settings used, number of trials, and settings of the search procedure if applicable.
As tuning of hyperparameters is typically performed using specific parts of the dataset, it is essential to note that any modeling decisions based on them automatically invalidate their use as \emph{test} data.

\paragraph{Models}
%Recently, Transformer-based models~\cite{vaswani2017attention, devlin2019bert, dosovitskiy2021image, chen2021decision} made a large impact on Deep Learning and NLP specifically. These models are large in memory and take significant compute to train them (e.g., BERT). To prevent retraining all these models, a recommendation is to save the models. In many of the Deep Learning libraries, e.g., \texttt{Keras}~\cite{chollet2015keras, gulli2017deep}, \texttt{PyTorch}~\cite{paszke2019pytorch}, TensorFlow~\cite{abadi2016tensorflow}, or \texttt{JAX}~\cite{jax2018github} allows for an option to save checkpoints which copies the weights of the model in that current state. These can then be redistributed and ran from that specific state. Models are too large in size to store in code repositories. A couple solutions could be making use of the ONNX format (\cref{app:resources}) or hosting it on the \texttt{HuggingFace} platform~\cite{wolf2020transformers} to name a few. 

%The recent surge of large Transformer-based models has had a large impact on DL and NLP~\cite{vaswani2017attention, devlin2019bert, dosovitskiy2021image, chen2021decision}. %These and many other 
Contemporary models (e.g.,~\citealp{vaswani2017attention, devlin2019bert, dosovitskiy2021image, chen2021decision}) have very large computational and memory footprints. To avoid retraining models, and more importantly, to allow for replicability, it is recommended to save and share model weights. This may face similar challenges as those of datasets (namely, large file sizes), but it remains an impactful consideration. In most cases, simply sharing the best or most interesting model could suffice. It should be emphasized that distributing model weights should always complement a well-documented repository as libraries and hosting sites might not be supported in the future.
%Note that in some cases, future distribution might fail, due to non-maintained repositories or broken hyperlinks.
% \todo{perhaps add that sharing only the best/most interesting model could be a solution?} %A fallback is strong documentation on the experiments for replicability.
%Then, strong experimental documentation in the research paper becomes a fallback to ensure some degree of replicability.

% \todo{elaborate on how we can't always share everything and explain what to do then.}

\paragraph{Model Evaluation} %Concerning models and tasks, 
The exact model and task evaluation procedure can differ significantly~\citep[e.g.][]{post-2018-call}.
It is important to either reference the exact evaluation script used (including parameters, citation, and version, if applicable) or include the evaluation script in the codebase. Moreover, to ease error or post-hoc analyses, we highly recommend saving model predictions whenever possible and making them available at publication~\cite{card2020with, gehrmann2022repairing}.
%formatting, documented Python's \texttt{pickle} modules, or Google's \texttt{protobuf} (see \cref{app:resources}) or at least in the same format as the annotated data.
% \todo{Shall we just use the format that the annotated data uses instead?, especially the last 2 I don't like}.
%This can aid future work and error analysis. %without having to re-implement or rerun experiments and help uncover flawed predictions.

% \paragraph{In the Paper} Specifically for the paper regarding the trend of replicability, we expect a report on the computing infrastructure used. At minimum, the type of Central Processing Unit (CPU) and the Graphics Processing Unit (GPU) are reported. This is for indicating the amount of compute necessary for the project, but also preventing reproducibility issues due to the non-deterministic nature of the GPU ~\citep{jean2019issues, wei2020leaky}. 

%Given the standard nowadays of using GPUs in plenum, they need energy to perform their task. Therefore, it is also important to report the energy usage of the model, and their emission~\cite{strubell-etal-2019-energy, schwartz2020green, henderson2020towards}. This is then to reconsider the amount of heavy compute necessary for large models.

\paragraph{Model Cards} Apart from quantitative evaluation and optimal hyperparameters,
%from the qualitative side of models, \citet{mitchell2019model} propose 
~\citet{mitchell2019model} propose model cards: A type of standardized documentation, as a step towards responsible ML and AI technology, accompanying trained ML models that provide benchmarked evaluation in a variety of conditions, across different cultural, demographic, or phenotypic and intersectional groups that are relevant to the intended application domains. They can be reported in the paper or project, and can help to collect important information for reproducibility, such as preprocessing and evaluation results. We refer to~\citet{mitchell2019model,menon2020pulse} for examples of model cards.

\begin{tcolorbox}[
    title={\centering\textcolor{black}{Best Practices: \textbf{Codebase \& Models}}}, 
    colback=white,
    colframe=red!25!green!70!blue!55, enlarge top by=5pt,
    left=2pt, right=8pt,
    boxsep=2pt,
    coltext=black
]
  \begin{itemize}[leftmargin=12pt]
      \small
      \itemsep0em
    %   \item[$\diamond$] Publish a code repository with detailed documentation including licensing to distribute code for replicability;
      \item[$\diamond$] Publish a code repository with documentation and licensing to distribute for replicability;
      \item[$\diamond$] Report all details about hyperparameter search and model training;
      \item[$\diamond$] Specify the hyperparameters for replicability;
      \item[$\diamond$] Publish model predictions and evaluation scripts.;
      \item[$\star$] Use model cards;
      \item[$\star$] Publish models; 
  \end{itemize}
\end{tcolorbox}

\section{Experiments \& Analysis}\label{sec:experiments-analysis}

Experiments and their analyses constitute the core of most scientific works, and empirical evidence is valued especially highly in ML research \citep{birhane2021values}. 
%As such, special care should be put into designing and executing them. 
%We outlined in the introduction how issues with replicability and significance of results in the ML literature have been raised by several authors \citep{gundersen2018state,henderson2018deep,narang2021transformer,schmidt2021descending}. 
Therefore, we discuss the most common issues and counter-strategies at different stages of an experiment.

%\subsection{Experimental Execution}\label{sec:experimental-execution}

%\paragraph{Data Splits} Considerations begin with having pre-defined splits of the dataset which are unseen during the final evaluation as well as intermediate tuning. Assuming that the overall dataset adheres to the aforementioned standards (\cref{sec:data}), these splits must also be conducted in a way as not to compromise them (e.g., by splitting based on a subpopulation). Recent surveys indicate that the way data are split has substantial impacts on later conclusions and that even test splits which have existed for a longer period of time can be prone being over-fit on, as more and more people aim to beat this particular benchmark \citep{gorman2019}.

\paragraph{Model Training} For model training, it is advisable to set a random seed for replicability, and train multiple initializations per model in order to obtain a sufficient sample size for later statistical tests. 
%Commonly used values are three to five runs, however this
The number of runs should be adapted based on the observed variance: Using for instance bootstrap power analysis, existing model scores are raised by a constant compared to the original sample using a significance test in a bootstrapping procedure \citep{yuan2003bootstrap,tuffery2011data, henderson2018deep}. If the percentage of significant results is low, we should collect more scores.\footnote{The resulting tensions with modern DL hardware requirements are discussed in \cref{sec:discussion}.}
%We are aware that this poses some tension with the hardware requirements of many modern DL architectures, which is why we dedicate part of the discussion in \cref{sec:discussion} to this question.} 
\citet{bouthillier2021accounting} further recommend to vary as many sources of randomness in the training procedure as possible (i.e., data shuffling, data splits etc.) to obtain a closer approximation of the true model performance. Nevertheless, any drawn conclusion are still surrounded by a degree of statistical uncertainty, which can be combated by the use of statistical hypothesis testing.

%Nevertheless, the question of what conclusions can be drawn from these outcomes can be harder than it might superficially seem, precisely due to the mentioned sources of statistical uncertainty. A common solution is the use of statistical hypothesis testing, which we portray here along with criticisms and alternatives.

%\subsection{Experimental Analysis}\label{sec:experimnental-analysis}

%\footnote{We also provide more references in .} Furthermore, \citet{azer2020not} provide a ready-made Python package for a number of significance tests and \citet{ulmer2022deep} an implementation of the test by \citet{dror2019deep} and more, which is tailored specifically towards neural networks (see \cref{app:resources}). \citet{agarwal2021deep} provide a test framework that is adapted for deep reinforcement learning. With the necessary tools at hand, we can now return to carefully answer the original research questions. %But even then, conclusions should be drawn carefully, as the methods listed previously all operate under a degree of statistical uncertainty.
%The choice can depend on: The variable type and distribution of test scores and whether it is known and the amount of observations. We give a few rules of thumb:

\paragraph{Significance Testing} 
%Especially with 
Using deep neural networks,  a number of (stochastic) factors such as the random seed \citep{dror2019deep} or even the choice of hardware~\citep{yang-etal-2018-design} or framework \citep{leventi2022deep} can influence performance and need to be taken into account. 
%As such, multiple factors have to be taken into account when drawing conclusions from experimental results. 
First of all, the size of the dataset should support sufficiently powered statistical analyses (see~\cref{sec:data}).
Secondly, an appropriate significance test should be chosen. We give a few rules of thumb based on \citet{dror2018hitchhiker}: When the distribution of scores is known, for instance a normal distribution for the Student's t-test, a \emph{parametric} test should be chosen. Parametric tests are designed with a specific distribution for the test statistic in mind, and have strong statistical power (i.e.\@ a lower Type II error). The underlying assumptions can sometimes be hard to verify (see \citealp{dror2018hitchhiker} §3.1), thus when in doubt \emph{non-parametric} tests can be used. This category features tests like the Bootstrap, employed in case of a small sample size, or the Wilcoxon signed-rank test \citep{wilcoxon1992individual}, when plenty observations are available. Depending on the application, the usage of specialized tests might furthermore be desirable \citep{dror2019deep, agarwal2021deep}. We also want to draw attention to the fact that comparisons between multiple models and/or datasets, \emph{require} an adjustment of the confidence level, for instance using the Bonferroni correction \citep{bonferroni1936teoria}, which is a safe and conservative choice and easily implemented for most tests \citep{dror2017replicability, ulmer2022deep}. \citet{azer2020not} provide a guide on how to adequately word insights when a statistical test was used, and \citet{greenland2016statistical} list common pitfalls and misinterpretations of results. 
Due to spatial constraints, we here refer to \cref{app:experimental-analysis} for a number of easy-to-use software packages and further reading on the topic. %\citet{dror2018hitchhiker,raschka2018model} for a general introduction to the topic and \citet{azer2020not} for an overview over Bayesian significance tests. In \cref{app:experimental-analysis}, we also list a number of resources, such as Bayesian significance tests by \citet{azer2020not}, an implementation of the test by \citet{dror2019deep} by \citet{ulmer2022deep} and a test framework that is adapted for deep reinforcement learning by \citet{agarwal2021deep}. With the necessary tools at hand, we can now return to carefully answer the original research questions.

\paragraph{Critiques \& Alternatives} Although statistical hypothesis testing is an established tool in many disciplines, its (mis)use has received criticism for decades \citep{berger1987testing, demvsar2008appropriateness, ziliak2008cult}. For instance, \citet{wasserstein2019moving} criticize the $p$-value as reinforcing publication bias through the dichotomy of ``significant'' and ``not significant'', i.e., by favoring positive results \citep{locascio2017results}. Instead, \citet{wasserstein2019moving} propose to report it as a continuous value and with the appropriate scepticism.\footnote{Or, as \citet{wasserstein2019moving} note: ``\emph{statistically significant}---don't say it and don't use it''.} In addition to statistical significance, another approach advocates for reporting \emph{effect size} \citep{berger1987testing, lin2013research}, so for instance the mean difference, or the absolute or relative gain in performance for a model compared to a baseline. The effect size can be modeled using Bayesian analysis \citep{kruschke2013bayesian, benavoli2017time}, which better fit the uncertainty surrounding experimental results, but requires the specification of a plausible statistical model producing the observations
\footnote{Here, we are \emph{not} referring to a neural network, but instead to a process generating experimental observations, specifying a prior and likelihood for model scores. Conclusions are drawn from the posterior distribution over parameters of interest (e.g., the mean performance), as demonstrated by \citet{benavoli2017time}.} 
and potentially the usage of Markov Chain Monte Carlo sampling \citep{brooks2011handbook, gelmanbayesian}. \citet{benavoli2017time} give a tutorial for applications to ML and supply an implementation of their proposed methods in a software package (see \cref{app:resources}) and guidelines for reporting details are given by \citet{kruschke2021bayesian}, including for instance the choice of model and priors. 

%\paragraph{Reporting Results} Lastly, report the number of runs/random seeds used, and, if appropriate, the significance threshold or confidence level of the statistical test or the underlying Bayesian model and chosen priors. Report all scores using mean and standard deviation and report $p$-values or comparable quantities as continuous values, not binary decisions. Using those results, evaluate the evidence for and against your initial hypotheses.

%\paragraph{Saving Predictions} The results of the model could be saved as well, to prevent inference time of the model. The format here is free, but recommendations are to save the model outputs the same as the model inputs. For example, if there is a tab-separated file consisting of a input and a target, the resulting test set should have the same format as \texttt{input / ID [tab] target}. 

\begin{tcolorbox}[
    title={\centering\textcolor{black}{Best Practices: \textbf{Experiments \& Analysis}}}, 
    colback=white,
    colframe=red!25!green!20!blue!25, 
    left=2pt, right=8pt, enlarge top by=5pt,
    boxsep=2pt,
    coltext=black
]
   \begin{itemize}[leftmargin=12pt]
      \small
      \itemsep0em
       \item[$\diamond$] Report mean \& standard dev.\ over multiple runs;
       \item[$\diamond$] Perform significance testing or Bayesian analysis and motivate your choice of method;
       %\item[$\star$] Save model predictions
       \item[$\diamond$] Carefully reflect on the amount of evidence regarding your initial hypotheses.
   \end{itemize}
\end{tcolorbox}

\section{Publication}
\label{sec:publication}

%Subsequent to all the prior consideration, the publication step of a research project allows the findings to be spread across the scientific community.
In this section, we discuss some additional trends in the DL field that researchers should consider when publishing their work, even though they might not directly be related to reproducibility \& replicability.
% As argued in previous sections, when publishing a paper, researchers should take care to describe all the details related to data, models, experiments and analysis.
% In this section, we discuss some additional trends in the AI field that researchers should consider as well in their work.

\paragraph{Citation Control} While frequently, researchers cite non-archival versions of papers, 
%without noticing that the paper has been published already. The
the published version of a paper is peer-reviewed, increasing the probability that any mistakes or ambiguities have been resolved.
In \cref{app:resources}, we suggest tools to verify the version of any cited papers.

\paragraph{Hardware Requirements} The paper should report the computing infrastructure used. At minimum, the specifics about the CPU and GPU. This is for indicating the amount of compute necessary for the project, but also for the sake of replicability issues due to the non-deterministic nature of the GPU~\citep{jean2019issues, wei2020leaky}.
Moreover, \citet{dodge-etal-2019-show} demonstrate that test performance scores alone are insufficient for claiming the dominance of a model over another, and argue for reporting additional performance details on validation data as a function of computation budget, which can also estimate the amount of computation required to obtain a given accuracy.

\paragraph{Environmental Impact}
The growth of computational resources required for DL over the last decade has led to financial and carbon footprint discussions in the AI community.
\citet{schwartz2020green} introduce the distinction between \emph{Red AI}---AI research that seek to obtain state-of-the-art results through the use of massive computational power---and \emph{Green AI}---AI research that yields novel results without increasing computational cost.
In the paper the authors propose to add \emph{efficiency} as an evaluation criterion alongside accuracy measures. 
\citet{hershcovich2022towards} advocate for the usage of a \emph{climate performance model card}, in which energy and emission statistics are being detailed.
\citet{strubell-etal-2019-energy} approximate financial and environmental costs of training a variety of models (e.g., BERT, GPT-2). %\footnote{\cref{app:resources} includes a tool for CO\textsubscript{2} estimation of computational models.}
In conclusion, to reduce costs and improve equity, they propose (1) \emph{Reporting training time and sensitivity to hyperparameters}, (2) \emph{Equitable access to computation resources}, and (3) \emph{Prioritizing computationally efficient hardware and algorithms} (\cref{app:resources} includes a tool for CO\textsubscript{2} estimation of computational models).

\paragraph{Social Impact}
The widespread of DL studies and their increasing use of human-produced data (e.g., from social media and personal devices) means the outcome of experiments and applications have direct effects on the lives of individuals. %~\citet{waseem2021disembodied} argue that 
Addressing and mitigating biases in ML is near-impossible as subjectivity is inescapable and thus converging in a universal truth may further harm already marginalized social groups \cite{waseem2021disembodied,parmar2022don}.
As a follow-up, \citealp{waseem2021disembodied} argue for a reflection on the consequences the imaginary objectivity of ML has on political choices. \citet{hovy-spruit-2016-social} analyze and discuss the social impact research may have beyond the more explored privacy issues.
They make an ethical analysis on social justice, i.e., equal opportunities for individuals and groups, and underline three problems of the mutual relationship between language, society and individuals: exclusion, over-generalization and overexposure.
%Overall, while it seems undeniable that many DL studies could potentially be harmful to individuals and minorities, only $<$ 1.3\% of all papers published since 2016 in the ACL Anthology have had the study approved by an ethical review board \citep{santy-etal-2021-use}.

%%%%%%%%%%%%%%%%%% Removed sentence/citation. It already seems controversial for too many ppl :')%%%%%%%%%%%%%%%%%%%

% The wide-spread of NLP studies and its increasing use in social media platforms, mean the outcome of NLP experiments and applications to have a direct effect on the lives of individual users. \citet{hovy-spruit-2016-social} state the problem and discuss the social impact NLP may have beyond the more explored privacy issues.
% They make an ethical analysis of NLP on social justice, i.e.\ equal opportunities for individuals and groups, and underline three problems of the mutual relationship between language, society and individuals: exclusion, over-generalization and overexposure.

% \citet{waseem2021disembodied} analyze the phenomena from a more practical ML point of view.
% They argue that addressing and mitigating biases in ML is near-impossible as subjectivity is inescapable and thus converging it in a universal truth may further harm already marginalized social groups.
% As a follow-up, the authors argue for a reflection on the consequences the imaginary objectivity of ML has on political choices.

\paragraph{Ethical Considerations}
There has been effort on the development of concrete ethical guidelines for researchers within the ACM Code of Ethics and Professional Conduct~\citep{acm-code}.
The Code lists seven principles stating how fundamental ethical principles apply to the conduct of a computing professional (like DL and NLP practitioners) and is based on two main ideas: computing professionals' actions change the world and the public good is always the primary consideration.
\citet{mohammad2021ethics} discusses the importance of going beyond individual models and datasets, back to the ethics of the task itself. As a practical recommendation, he presents \emph{Ethics Sheets for AI Tasks} as tools to document ethical considerations \emph{before} building datasets and developing systems.
In addition, researchers are invited to collect the ethical considerations of the paper in a cohesive narrative, and elaborate them in a paragraph, usually in the Introduction/Motivation, Data, Evaluation, Error Analysis or Limitations section \citep{panelMohammad, bias_statement}.

\definecolor{betterred}{HTML}{DC8385}
\begin{tcolorbox}[
    title={\centering\textcolor{black}{Best Practices: \textbf{Publication}}}, 
    colback=white,
    colframe=betterred, 
    left=2pt, right=8pt, enlarge top by=5pt,
    boxsep=2pt,
    coltext=black
]
   \begin{itemize}[leftmargin=12pt]
  \small
      \itemsep0em
          \item[$\diamond$] Avoid citing pre-prints (if applicable);
       \item[$\diamond$] Describe the computational requirements;
       \item[$\diamond$] Consider the potential ethical \& social impact;%todo{I am also afraid of this one. For me this would mean, writing a paragraph, and copy it over to the rest of the papers I will write in my life, not super helpful}
       \item[$\star$] Consider the environmental impact and prioritize computational efficiency;
       \item[$\star$] Include an Ethics and/or Bias Statement.
   \end{itemize}
\end{tcolorbox}
%\todo{I would again disagree with your diamonds and stars, not all work should be on efficiency, also not all work needs an ethics statement (in fact, the minority needs them in my opinion). If you disagree with me, read the last paper you published, or the one you are working on now; do you follow these best practices yourself The first and third seem more important to me?}
\section{Discussion}\label{sec:discussion}

Since previous sections have emphasized the need to overhaul some experimental standards, we dedicate this last section to discuss some structural issues that might pose obstacles to this. 
%in a decidedly opinionated way.

% Hardware requirements vs. sample size
\paragraph{Compute Requirements} Specifically with regard to statistical significance in \cref{sec:experiments-analysis}, there is a stark tension between the hardware requirements of modern methods \citep{sevilla2022compute} and the computational budget of the average researcher. Only the best-funded research labs can afford the increasing computational costs to account for the statistical uncertainty of results and to reproduce prior works \citep{hooker2021hardware}. 
%Significance tests require many runs to produce reliable results: Neural network performance may fluctuate wildly,\footnote{E.g., BERT models with different seeds \citep{sellam2021multiberts} sometimes perform quite differently compared to the originally released model \citep{devlin2019bert}.} and thus pose daunting computational costs, which but the best-funded research labs can afford \citep{hooker2021hardware}. 
Under these circumstances, it becomes difficult to judge whether the results obtained via larger models and datasets \emph{actually} constitute substantial progress or just statistical flukes. At the same time, such experiments can create environmental concerns \citep{strubell-etal-2019-energy, schwartz2020green}.\footnote{E.g., GPT-3's training was estimated to have cost ca.\ 12M USD \citep{turner2020gpt3} or 188,702 kWh \citep{anthony2020carbontracker}.}
%and ethical nature  \citep{bender2021on}, while potentially reinforcing global inequalities and power structures \citep{mohamed2020decolonial}. 
The community must decide collectively whether these factors, including impeded reproducibility and weakened empirical evidence, constitute a worthy price for the knowledge obtained from training large neural networks.   

%The ``bitter lesson'' of Deep Learning \citep{sutton2019bitter}, i.e., the success of scaling data sets and architectures, has recently seemed to yield diminishing returns. For instance, the trend of increasingly large language models has provided decreasing benefits \citep{thompson2021deep}, while creating exuberant training costs of gargantuan models,\footnote{For instance, based on the compute requirements, GPT-3 was estimated to have cost around 12 million USD to train (\url{https://twitter.com/eturner303/status/1266264358771757057}), or 188701.92 kWh of energy \citep{anthony2020carbontracker}. In their work, the authors admit they discovered some train / test set overlap, but could not repeat the training due to cost \citep{brown2020language}.} beyond potential ethical \citep{bender2021on} and environmental concerns \citep{strubell-etal-2019-energy, schwartz2020green}, as well as reinforcing global inequalities and power structures \citep{mohamed2020decolonial}. The truth however is that it is doubtful whether we will be able to scale our way to Artificial General Intelligence. For instance, \citet{merrill2021provable} provided a formal argument on why large language models are unlikely to infer the meaning of expressions from raw text alone and even the largest models seem to not have made any progress on many of the aspects of human cognition missing in neural networks posed by \citet{lake2017building}. Nevertheless, scaling provides an incremental way to create publishable insights --- if the necessary resources are available.
 
% Incentive structures in ML publishing, also look at other disciplines
\paragraph{Incentives in Publishing} As demonstrated by \cref{fig:development-nlp}, NLP has gained traction as an empirical field of research. At such a point, more rigorous standards are necessary to maintain high levels of scholarship. Unfortunately, we see this process lagging behind, illustrated by repeated calls for improvement~\citep{gundersen2018state, narang2021transformer}. Why is that so?
We speculate that the reason for many of these problems are caused by adverse incentives set by the current publishing environment: As the career of researchers hinges on their publications and more rigorous experimental standards are often not required to get published, reproducing and reproducible works are not rewarded.
Instead, actors are tempted to ``rig the benchmark lottery'' \citep{dehghani2021benchmark}, since achieving state-of-the-art results remains important for publishing \citep{birhane2021values}.
 As of now, better experimental standards often do not increase the acceptance probability: The more details are provided for replicability purposes, the more potential points of criticism are exposed to reviewers. 
% Under these circumstances, the quality of published works decreases, replication suffers and peer review load increases. 
This state of affairs might still seem like progress to some, but \citet{chu2021slowed} show how an increased amount of papers actually leads to \emph{slowed} progress in a field, making it harder for new, promising ideas to break through. Furthermore, \citet{raff2022does} shows that reproducible work can actually have a positive impact on  a paper's citation rates, and thus should be more embraced.% by the community.

\paragraph{Culture Change} How can we change this trend? \textbf{As researchers}, we can start implementing the recommendations in this work in order to drive bottom-up change and reach a critical mass \citep{centola2018experimental}. \textbf{As reviewers}, we can shift focus from results to more rigorous methodologies \citep{rogers2021how}, and allow more critiques and reproductions of past works and meta-reviews to be published \citep{birhane2021values,lampinen2021publishing}. \textbf{As a community}, we can change the incentives around research and experiment with new initiatives. \citet{rogers2020what} and \citet{su2021you} give recommendations on how to improve the peer-review process by better paper-reviewer matching and paper scoring.
%\footnote{\citet{faggion2016improving} for instance has argued to also make the reviewing process itself more transparent. We can see developments in this direction in the adoption of OpenReview.} 
Other attempts are currently undertaken to encourage reproduction of past works.\footnote{See for instance the reproducibility certication of the TMLR journal \citep{tmlr2022guidelines} or NAACL 2022 reproducibility badges \citep{naacl2022reproducibility}.} Other ideas change the publishing process more fundamentally, for instance by splitting it into two steps: The first part, where authors are judged solely on the merit of their research question and methodology; and the second one, during which the analysis of their results is evaluated \citep{locascio2017results}. 
%This aims to reduce publication bias and puts more scrutiny on the experimental methodology. 
In a similar vein, \citet{miltenburg2021preregistering} recommend a procedure similar to clinical studies, where whole research projects are pre-registered, i.e., specifying the parameters of research before carrying out any experiments \citep{nosek2018preregistration}. The implications of these ideas are not only positive, however, as a slowing rate of publishing might disadvantage junior researchers \citep{chu2021slowed}.

\section{Conclusion}

Being able to (re-)produce empirical findings is critical for scientific progress, particularly in fast-growing fields like NLP~\cite{manning-2015-last}. To reduce the risks of a reproducibility crisis and unreliable research findings~\cite{Ioannidis2005}, experimental rigor is imperative. 
Being aware of possible harmful implications and to avoid them is therefore important. Every step carries possible biases~\cite{hovyshrimai21, waseem2021disembodied}.
%While necessarily incomplete, 
This paper aims at providing a toolbox of actionable recommendations \emph{for each research step}, and a reflection and summary of the ongoing broader discussion. With concrete best practices to raise awareness and call for uptake, we hope to aid researchers in their empirical endeavors.

\section*{Limitations}
This work comes with two main limitations: On the one hand, it can only take a snapshot of an ongoing discussion. On the other hand, this work was aimed to primarily serve the NLP community, although other disciplines using DL might also profit from these guidelines. With these limitations in mind, we invite members of the community to contribute to our open-source repository.

\section*{Ethics Statement}

We do not foresee any immediate negative ethical consequences in lieu with our work. 
\section*{Acknowledgments}

We would like to thank Giovanni Cin\`a, Rotem Dror, Miryam de Lhoneux, and Tanja Samard\v{z}i\'{c} for their feedback on this draft. Furthermore, we would like to express our gratitude to the NLPnorth group in general for frequent discussions and feedback on this work.
MZ and BP are supported by the Independent Research Fund Denmark (DFF) grant 9131-00019B. EB, MME, and BP are supported by the Independent Research Fund Denmark (DFF) Sapere Aude grant 9063-00077B. BP is supported by the ERC Consolidator Grant DIALECT  101043235.

% BP: starting to add people for this section 
% Tanja Samardžić
% Rotem Dror
% Giovanni Cin\`a
% Miryam de Lhoneux
% NLPNorth group in general

\bibliography{anthology,custom}

\begin{thebibliography}{163}
\expandafter\ifx\csname natexlab\endcsname\relax\def\natexlab#1{#1}\fi

\bibitem[{Agarwal et~al.(2021)Agarwal, Schwarzer, Castro, Courville, and
  Bellemare}]{agarwal2021deep}
Rishabh Agarwal, Max Schwarzer, Pablo~Samuel Castro, Aaron~C Courville, and
  Marc Bellemare. 2021.
\newblock Deep reinforcement learning at the edge of the statistical precipice.
\newblock \emph{Advances in Neural Information Processing Systems}, 34.

\bibitem[{Anthony et~al.(2020)Anthony, Kanding, and
  Selvan}]{anthony2020carbontracker}
Lasse F~Wolff Anthony, Benjamin Kanding, and Raghavendra Selvan. 2020.
\newblock Carbontracker: Tracking and predicting the carbon footprint of
  training deep learning models.
\newblock \emph{arXiv preprint arXiv:2007.03051}.

\bibitem[{Apidianaki et~al.(2018)Apidianaki, Mohammad, May, Shutova, Bethard,
  and Carpuat}]{apidianaki2018proceedings}
Marianna Apidianaki, Saif Mohammad, Jonathan May, Ekaterina Shutova, Steven
  Bethard, and Marine Carpuat. 2018.
\newblock Proceedings of the 12th international workshop on semantic
  evaluation.
\newblock In \emph{Proceedings of The 12th International Workshop on Semantic
  Evaluation}.

\bibitem[{{Association for Computational
  Linguistics}(2022)}]{naacl2022reproducibility}
{Association for Computational Linguistics}. 2022.
\newblock Reproducibility criteria.
\newblock \url{https://2022.naacl.org/calls/papers/#reproducibility-criteria}.
\newblock Accessed: 2022-02-09.

\bibitem[{{Association for Computing Machinery}(2022)}]{acm-code}
{Association for Computing Machinery}. 2022.
\newblock Acm code of ethics and professional conduct.
\newblock \url{https://www.acm.org/code-of-ethics}.
\newblock Accessed: 2022-02-10.

\bibitem[{Azer et~al.(2020)Azer, Khashabi, Sabharwal, and Roth}]{azer2020not}
Erfan~Sadeqi Azer, Daniel Khashabi, Ashish Sabharwal, and Dan Roth. 2020.
\newblock \href {https://doi.org/10.18653/v1/2020.acl-main.506} {Not all claims
  are created equal: Choosing the right statistical approach to assess
  hypotheses}.
\newblock In \emph{Proceedings of the 58th Annual Meeting of the Association
  for Computational Linguistics, {ACL} 2020, Online, July 5-10, 2020}, pages
  5715--5725. Association for Computational Linguistics.

\bibitem[{Basile et~al.(2021)Basile, Fell, Fornaciari, Hovy, Paun, Plank,
  Poesio, Uma et~al.}]{basile2021we}
Valerio Basile, Michael Fell, Tommaso Fornaciari, Dirk Hovy, Silviu Paun,
  Barbara Plank, Massimo Poesio, Alexandra Uma, et~al. 2021.
\newblock We need to consider disagreement in evaluation.
\newblock In \emph{1st Workshop on Benchmarking: Past, Present and Future},
  pages 15--21. Association for Computational Linguistics.

\bibitem[{Bassignana and Plank(2022)}]{elisaspaper}
Elisa Bassignana and Barbara Plank. 2022.
\newblock Cross{RE}: A {C}ross-{D}omain {D}ataset for {R}elation {E}xtraction.
\newblock In \emph{Findings of the Association for Computational Linguistics:
  EMNLP 2022}. Association for Computational Linguistics.

\bibitem[{Basta et~al.(2019)Basta, Costa-juss{\`a}, and
  Casas}]{basta2019evaluating}
Christine Basta, Marta~R Costa-juss{\`a}, and Noe Casas. 2019.
\newblock Evaluating the underlying gender bias in contextualized word
  embeddings.
\newblock In \emph{Proceedings of the First Workshop on Gender Bias in Natural
  Language Processing}, pages 33--39.

\bibitem[{Belz et~al.(2020)Belz, Agarwal, Shimorina, and
  Reiter}]{belz-etal-2020-reprogen}
Anya Belz, Shubham Agarwal, Anastasia Shimorina, and Ehud Reiter. 2020.
\newblock \href {https://aclanthology.org/2020.inlg-1.29} {{R}epro{G}en:
  Proposal for a shared task on reproducibility of human evaluations in {NLG}}.
\newblock In \emph{Proceedings of the 13th International Conference on Natural
  Language Generation}, pages 232--236, Dublin, Ireland. Association for
  Computational Linguistics.

\bibitem[{Belz et~al.(2021)Belz, Agarwal, Shimorina, and
  Reiter}]{belz2021systematic}
Anya Belz, Shubham Agarwal, Anastasia Shimorina, and Ehud Reiter. 2021.
\newblock \href {https://aclanthology.org/2021.eacl-main.29/} {A systematic
  review of reproducibility research in natural language processing}.
\newblock In \emph{Proceedings of the 16th Conference of the European Chapter
  of the Association for Computational Linguistics: Main Volume, {EACL} 2021,
  Online, April 19 - 23, 2021}, pages 381--393. Association for Computational
  Linguistics.

\bibitem[{Benavoli et~al.(2017)Benavoli, Corani, Demsar, and
  Zaffalon}]{benavoli2017time}
Alessio Benavoli, Giorgio Corani, Janez Demsar, and Marco Zaffalon. 2017.
\newblock \href {http://jmlr.org/papers/v18/16-305.html} {Time for a change: a
  tutorial for comparing multiple classifiers through bayesian analysis}.
\newblock \emph{J. Mach. Learn. Res.}, 18:77:1--77:36.

\bibitem[{Benavoli et~al.(2014)Benavoli, Corani, Mangili, Zaffalon, and
  Ruggeri}]{benavoli2014bayesian}
Alessio Benavoli, Giorgio Corani, Francesca Mangili, Marco Zaffalon, and
  Fabrizio Ruggeri. 2014.
\newblock A bayesian wilcoxon signed-rank test based on the dirichlet process.
\newblock In \emph{International conference on machine learning}, pages
  1026--1034. PMLR.

\bibitem[{Bender and Friedman(2018)}]{bender2018data}
Emily~M. Bender and Batya Friedman. 2018.
\newblock \href {https://doi.org/10.1162/tacla00041} {Data statements for
  natural language processing: Toward mitigating system bias and enabling
  better science}.
\newblock \emph{Transactions of the Association for Computational Linguistics},
  6:58--04.

\bibitem[{Bender et~al.(2021)Bender, Gebru, McMillan-Major, and
  Shmitchell}]{bender2021dangers}
Emily~M Bender, Timnit Gebru, Angelina McMillan-Major, and Shmargaret
  Shmitchell. 2021.
\newblock On the dangers of stochastic parrots: Can language models be too big?
\newblock In \emph{Proceedings of the 2021 ACM Conference on Fairness,
  Accountability, and Transparency}, pages 610--623.

\bibitem[{Benjamin(2018)}]{benjamin2018}
Martin Benjamin. 2018.
\newblock Hard numbers: Language exclusion in computational linguistics and
  natural language processing.
\newblock In \emph{Proceedings of the LREC 2018 Workshop
  “CCURL2018--Sustaining Knowledge Diversity in the Digital Age}, pages
  13--18.

\bibitem[{Berger and Sellke(1987)}]{berger1987testing}
James~O Berger and Thomas Sellke. 1987.
\newblock Testing a point null hypothesis: The irreconcilability of p values
  and evidence.
\newblock \emph{Journal of the American statistical Association},
  82(397):112--122.

\bibitem[{Bergstra and Bengio(2012)}]{bergstra2012random}
James Bergstra and Yoshua Bengio. 2012.
\newblock Random search for hyper-parameter optimization.
\newblock \emph{Journal of machine learning research}, 13(2).

\bibitem[{Birhane et~al.(2021)Birhane, Kalluri, Card, Agnew, Dotan, and
  Bao}]{birhane2021values}
Abeba Birhane, Pratyusha Kalluri, Dallas Card, William Agnew, Ravit Dotan, and
  Michelle Bao. 2021.
\newblock The values encoded in machine learning research.
\newblock \emph{arXiv preprint arXiv:2106.15590}.

\bibitem[{Bischl et~al.(2021)Bischl, Binder, Lang, Pielok, Richter, Coors,
  Thomas, Ullmann, Becker, Boulesteix et~al.}]{bischl2021hyperparameter}
Bernd Bischl, Martin Binder, Michel Lang, Tobias Pielok, Jakob Richter, Stefan
  Coors, Janek Thomas, Theresa Ullmann, Marc Becker, Anne-Laure Boulesteix,
  et~al. 2021.
\newblock Hyperparameter optimization: Foundations, algorithms, best practices
  and open challenges.
\newblock \emph{arXiv preprint arXiv:2107.05847}.

\bibitem[{Bonferroni(1936)}]{bonferroni1936teoria}
Carlo Bonferroni. 1936.
\newblock Teoria statistica delle classi e calcolo delle probabilita.
\newblock \emph{Pubblicazioni del R Istituto Superiore di Scienze Economiche e
  Commericiali di Firenze}, 8:3--62.

\bibitem[{Bouthillier et~al.(2021)Bouthillier, Delaunay, Bronzi, Trofimov,
  Nichyporuk, Szeto, Mohammadi~Sepahvand, Raff, Madan, Voleti
  et~al.}]{bouthillier2021accounting}
Xavier Bouthillier, Pierre Delaunay, Mirko Bronzi, Assya Trofimov, Brennan
  Nichyporuk, Justin Szeto, Nazanin Mohammadi~Sepahvand, Edward Raff, Kanika
  Madan, Vikram Voleti, et~al. 2021.
\newblock Accounting for variance in machine learning benchmarks.
\newblock \emph{Proceedings of Machine Learning and Systems}, 3.

\bibitem[{Bowman and Dahl(2021)}]{bowman-dahl-2021-will}
Samuel~R. Bowman and George Dahl. 2021.
\newblock \href {https://doi.org/10.18653/v1/2021.naacl-main.385} {What will it
  take to fix benchmarking in natural language understanding?}
\newblock In \emph{Proceedings of the 2021 Conference of the North American
  Chapter of the Association for Computational Linguistics: Human Language
  Technologies}, pages 4843--4855, Online. Association for Computational
  Linguistics.

\bibitem[{Braggaar and van~der
  Goot(2021)}]{braggaar-van-der-goot-2021-challenges}
Anouck Braggaar and Rob van~der Goot. 2021.
\newblock \href {https://aclanthology.org/2021.adaptnlp-1.6} {Challenges in
  annotating and parsing spoken, code-switched, {F}risian-{D}utch data}.
\newblock In \emph{Proceedings of the Second Workshop on Domain Adaptation for
  NLP}, pages 50--58, Kyiv, Ukraine. Association for Computational Linguistics.

\bibitem[{Brooks et~al.(2011)Brooks, Gelman, Jones, and
  Meng}]{brooks2011handbook}
Steve Brooks, Andrew Gelman, Galin Jones, and Xiao-Li Meng. 2011.
\newblock \emph{Handbook of markov chain monte carlo}.
\newblock CRC press.

\bibitem[{Brown et~al.(2020)Brown, Mann, Ryder, Subbiah, Kaplan, Dhariwal,
  Neelakantan, Shyam, Sastry, Askell, Agarwal, Herbert-Voss, Krueger, Henighan,
  Child, Ramesh, Ziegler, Wu, Winter, Hesse, Chen, Sigler, Litwin, Gray, Chess,
  Clark, Berner, McCandlish, Radford, Sutskever, and Amodei}]{brown-gpt3}
Tom Brown, Benjamin Mann, Nick Ryder, Melanie Subbiah, Jared~D Kaplan, Prafulla
  Dhariwal, Arvind Neelakantan, Pranav Shyam, Girish Sastry, Amanda Askell,
  Sandhini Agarwal, Ariel Herbert-Voss, Gretchen Krueger, Tom Henighan, Rewon
  Child, Aditya Ramesh, Daniel Ziegler, Jeffrey Wu, Clemens Winter, Chris
  Hesse, Mark Chen, Eric Sigler, Mateusz Litwin, Scott Gray, Benjamin Chess,
  Jack Clark, Christopher Berner, Sam McCandlish, Alec Radford, Ilya Sutskever,
  and Dario Amodei. 2020.
\newblock \href
  {https://proceedings.neurips.cc/paper/2020/file/1457c0d6bfcb4967418bfb8ac142f64a-Paper.pdf}
  {Language models are few-shot learners}.
\newblock In \emph{Advances in Neural Information Processing Systems},
  volume~33, pages 1877--1901. Curran Associates, Inc.

\bibitem[{Card et~al.(2020)Card, Henderson, Khandelwal, Jia, Mahowald, and
  Jurafsky}]{card2020with}
Dallas Card, Peter Henderson, Urvashi Khandelwal, Robin Jia, Kyle Mahowald, and
  Dan Jurafsky. 2020.
\newblock \href {https://doi.org/10.18653/v1/2020.emnlp-main.745} {With little
  power comes great responsibility}.
\newblock In \emph{Proceedings of the 2020 Conference on Empirical Methods in
  Natural Language Processing, {EMNLP} 2020, Online, November 16-20, 2020},
  pages 9263--9274. Association for Computational Linguistics.

\bibitem[{Carroll(2019)}]{carroll2019beyond}
Sean~M Carroll. 2019.
\newblock Beyond falsifiability: Normal science in a multiverse.
\newblock \emph{Why trust a theory}, pages 300--314.

\bibitem[{Caswell et~al.(2020)Caswell, Breiner, van Esch, and
  Bapna}]{caswell-etal-2020-language}
Isaac Caswell, Theresa Breiner, Daan van Esch, and Ankur Bapna. 2020.
\newblock \href {https://doi.org/10.18653/v1/2020.coling-main.579} {Language
  {ID} in the wild: Unexpected challenges on the path to a thousand-language
  web text corpus}.
\newblock In \emph{Proceedings of the 28th International Conference on
  Computational Linguistics}, pages 6588--6608, Barcelona, Spain (Online).
  International Committee on Computational Linguistics.

\bibitem[{Caswell et~al.(2021)Caswell, Kreutzer, Wang, Wahab, van Esch,
  Ulzii-Orshikh, Tapo, Subramani, Sokolov, Sikasote, Setyawan, Sarin, Samb,
  Sagot, Rivera, Rios, Papadimitriou, Osei, Suárez, Orife, Ogueji, Niyongabo,
  Nguyen, Müller, Müller, Muhammad, Muhammad, Mnyakeni, Mirzakhalov,
  Matangira, Leong, Lawson, Kudugunta, Jernite, Jenny, Firat, Dossou, Dlamini,
  de~Silva, Çabuk Ballı, Biderman, Battisti, Baruwa, Bapna, Baljekar, Azime,
  Awokoya, Ataman, Ahia, Ahia, Agrawal, and Adeyemi}]{caswell2021quality}
Isaac Caswell, Julia Kreutzer, Lisa Wang, Ahsan Wahab, Daan van Esch,
  Nasanbayar Ulzii-Orshikh, Allahsera Tapo, Nishant Subramani, Artem Sokolov,
  Claytone Sikasote, Monang Setyawan, Supheakmungkol Sarin, Sokhar Samb,
  Benoît Sagot, Clara Rivera, Annette Rios, Isabel Papadimitriou, Salomey
  Osei, Pedro Javier~Ortiz Suárez, Iroro Orife, Kelechi Ogueji, Rubungo~Andre
  Niyongabo, Toan~Q. Nguyen, Mathias Müller, André Müller,
  Shamsuddeen~Hassan Muhammad, Nanda Muhammad, Ayanda Mnyakeni, Jamshidbek
  Mirzakhalov, Tapiwanashe Matangira, Colin Leong, Nze Lawson, Sneha Kudugunta,
  Yacine Jernite, Mathias Jenny, Orhan Firat, Bonaventure F.~P. Dossou, Sakhile
  Dlamini, Nisansa de~Silva, Sakine Çabuk Ballı, Stella Biderman, Alessia
  Battisti, Ahmed Baruwa, Ankur Bapna, Pallavi Baljekar, Israel~Abebe Azime,
  Ayodele Awokoya, Duygu Ataman, Orevaoghene Ahia, Oghenefego Ahia, Sweta
  Agrawal, and Mofetoluwa Adeyemi. 2021.
\newblock \href {http://arxiv.org/abs/2103.12028} {Quality at a glance: An
  audit of web-crawled multilingual datasets}.

\bibitem[{Centola et~al.(2018)Centola, Becker, Brackbill, and
  Baronchelli}]{centola2018experimental}
Damon Centola, Joshua Becker, Devon Brackbill, and Andrea Baronchelli. 2018.
\newblock Experimental evidence for tipping points in social convention.
\newblock \emph{Science}, 360(6393):1116--1119.

\bibitem[{Chen et~al.(2021)Chen, Lu, Rajeswaran, Lee, Grover, Laskin, Abbeel,
  Srinivas, and Mordatch}]{chen2021decision}
Lili Chen, Kevin Lu, Aravind Rajeswaran, Kimin Lee, Aditya Grover, Misha
  Laskin, Pieter Abbeel, Aravind Srinivas, and Igor Mordatch. 2021.
\newblock Decision transformer: Reinforcement learning via sequence modeling.
\newblock \emph{Advances in neural information processing systems}, 34.

\bibitem[{Chu and Evans(2021)}]{chu2021slowed}
Johan~SG Chu and James~A Evans. 2021.
\newblock Slowed canonical progress in large fields of science.
\newblock \emph{Proceedings of the National Academy of Sciences}, 118(41).

\bibitem[{Cohen et~al.(2018)Cohen, Xia, Zweigenbaum, Callahan, Hargraves, Goss,
  Ide, N{\'e}v{\'e}ol, Grouin, and Hunter}]{cohen2018three}
K~Bretonnel Cohen, Jingbo Xia, Pierre Zweigenbaum, Tiffany~J Callahan, Orin
  Hargraves, Foster Goss, Nancy Ide, Aur{\'e}lie N{\'e}v{\'e}ol, Cyril Grouin,
  and Lawrence~E Hunter. 2018.
\newblock Three dimensions of reproducibility in natural language processing.
\newblock In \emph{LREC... International Conference on Language Resources \&
  Evaluation:[proceedings]. International Conference on Language Resources and
  Evaluation}, volume 2018, page 156. NIH Public Access.

\bibitem[{Corani and Benavoli(2015)}]{corani2015bayesian}
Giorgio Corani and Alessio Benavoli. 2015.
\newblock A bayesian approach for comparing cross-validated algorithms on
  multiple data sets.
\newblock \emph{Machine Learning}, 100(2):285--304.

\bibitem[{Curth et~al.(2021)Curth, Svensson, Weatherall, and van~der
  Schaar}]{curth2021really}
Alicia Curth, David Svensson, James Weatherall, and Mihaela van~der Schaar.
  2021.
\newblock Really doing great at estimating cate? a critical look at ml
  benchmarking practices in treatment effect estimation.

\bibitem[{Curth and van~der
  Schaar(2021{\natexlab{a}})}]{curth2021nonparametric}
Alicia Curth and Mihaela van~der Schaar. 2021{\natexlab{a}}.
\newblock Nonparametric estimation of heterogeneous treatment effects: From
  theory to learning algorithms.
\newblock In \emph{Proceedings of the 24th International Conference on
  Artificial Intelligence and Statistics (AISTATS)}. PMLR.

\bibitem[{Curth and van~der Schaar(2021{\natexlab{b}})}]{curth2021inductive}
Alicia Curth and Mihaela van~der Schaar. 2021{\natexlab{b}}.
\newblock On inductive biases for heterogeneous treatment effect estimation.

\bibitem[{D'Amour et~al.(2020)D'Amour, Heller, Moldovan, Adlam, Alipanahi,
  Beutel, Chen, Deaton, Eisenstein, Hoffman, Hormozdiari, Houlsby, Hou, Jerfel,
  Karthikesalingam, Lucic, Ma, McLean, Mincu, Mitani, Montanari, Nado,
  Natarajan, Nielson, Osborne, Raman, Ramasamy, Sayres, Schrouff, Seneviratne,
  Sequeira, Suresh, Veitch, Vladymyrov, Wang, Webster, Yadlowsky, Yun, Zhai,
  and Sculley}]{amour2011underspecification}
Alexander D'Amour, Katherine~A. Heller, Dan Moldovan, Ben Adlam, Babak
  Alipanahi, Alex Beutel, Christina Chen, Jonathan Deaton, Jacob Eisenstein,
  Matthew~D. Hoffman, Farhad Hormozdiari, Neil Houlsby, Shaobo Hou, Ghassen
  Jerfel, Alan Karthikesalingam, Mario Lucic, Yi{-}An Ma, Cory~Y. McLean, Diana
  Mincu, Akinori Mitani, Andrea Montanari, Zachary Nado, Vivek Natarajan,
  Christopher Nielson, Thomas~F. Osborne, Rajiv Raman, Kim Ramasamy, Rory
  Sayres, Jessica Schrouff, Martin Seneviratne, Shannon Sequeira, Harini
  Suresh, Victor Veitch, Max Vladymyrov, Xuezhi Wang, Kellie Webster, Steve
  Yadlowsky, Taedong Yun, Xiaohua Zhai, and D.~Sculley. 2020.
\newblock \href {http://arxiv.org/abs/2011.03395} {Underspecification presents
  challenges for credibility in modern machine learning}.
\newblock \emph{CoRR}, abs/2011.03395.

\bibitem[{Dehghani et~al.(2021)Dehghani, Tay, Gritsenko, Zhao, Houlsby, Diaz,
  Metzler, and Vinyals}]{dehghani2021benchmark}
Mostafa Dehghani, Yi~Tay, Alexey~A Gritsenko, Zhe Zhao, Neil Houlsby, Fernando
  Diaz, Donald Metzler, and Oriol Vinyals. 2021.
\newblock The benchmark lottery.

\bibitem[{Dem{\v{s}}ar(2008)}]{demvsar2008appropriateness}
Janez Dem{\v{s}}ar. 2008.
\newblock On the appropriateness of statistical tests in machine learning.
\newblock In \emph{Workshop on Evaluation Methods for Machine Learning in
  conjunction with ICML}, page~65. Citeseer.

\bibitem[{Devlin et~al.(2019)Devlin, Chang, Lee, and
  Toutanova}]{devlin2019bert}
Jacob Devlin, Ming-Wei Chang, Kenton Lee, and Kristina Toutanova. 2019.
\newblock Bert: Pre-training of deep bidirectional transformers for language
  understanding.
\newblock In \emph{Proceedings of the 2019 Conference of the North American
  Chapter of the Association for Computational Linguistics: Human Language
  Technologies, Volume 1 (Long and Short Papers)}, pages 4171--4186.

\bibitem[{Dodge et~al.(2019)Dodge, Gururangan, Card, Schwartz, and
  Smith}]{dodge-etal-2019-show}
Jesse Dodge, Suchin Gururangan, Dallas Card, Roy Schwartz, and Noah~A. Smith.
  2019.
\newblock \href {https://doi.org/10.18653/v1/D19-1224} {Show your work:
  Improved reporting of experimental results}.
\newblock In \emph{Proceedings of the 2019 Conference on Empirical Methods in
  Natural Language Processing and the 9th International Joint Conference on
  Natural Language Processing (EMNLP-IJCNLP)}, pages 2185--2194, Hong Kong,
  China. Association for Computational Linguistics.

\bibitem[{Dodge and Smith(2020)}]{emnlp_rep}
Jesse Dodge and Noah~A. Smith. 2020.
\newblock Reproducibility at emnlp 2020.
\newblock \url{https://2020.emnlp.org/blog/2020-05-20-reproducibility}.

\bibitem[{Dosovitskiy et~al.(2021)Dosovitskiy, Beyer, Kolesnikov, Weissenborn,
  Zhai, Unterthiner, Dehghani, Minderer, Heigold, Gelly, Uszkoreit, and
  Houlsby}]{dosovitskiy2021image}
Alexey Dosovitskiy, Lucas Beyer, Alexander Kolesnikov, Dirk Weissenborn,
  Xiaohua Zhai, Thomas Unterthiner, Mostafa Dehghani, Matthias Minderer, Georg
  Heigold, Sylvain Gelly, Jakob Uszkoreit, and Neil Houlsby. 2021.
\newblock An image is worth 16x16 words: Transformers for image recognition at
  scale.
\newblock In \emph{9th International Conference on Learning Representations,
  {ICLR} 2021, Virtual Event, Austria, May 3-7, 2021}. OpenReview.net.

\bibitem[{Dror et~al.(2017)Dror, Baumer, Bogomolov, and
  Reichart}]{dror2017replicability}
Rotem Dror, Gili Baumer, Marina Bogomolov, and Roi Reichart. 2017.
\newblock \href {https://transacl.org/ojs/index.php/tacl/article/view/1241}
  {Replicability analysis for natural language processing: Testing significance
  with multiple datasets}.
\newblock \emph{Trans. Assoc. Comput. Linguistics}, 5:471--486.

\bibitem[{Dror et~al.(2018)Dror, Baumer, Shlomov, and
  Reichart}]{dror2018hitchhiker}
Rotem Dror, Gili Baumer, Segev Shlomov, and Roi Reichart. 2018.
\newblock \href {https://doi.org/10.18653/v1/P18-1128} {The hitchhiker's guide
  to testing statistical significance in natural language processing}.
\newblock In \emph{Proceedings of the 56th Annual Meeting of the Association
  for Computational Linguistics, {ACL} 2018, Melbourne, Australia, July 15-20,
  2018, Volume 1: Long Papers}, pages 1383--1392. Association for Computational
  Linguistics.

\bibitem[{Dror et~al.(2020)Dror, Peled-Cohen, Shlomov, and
  Reichart}]{dror2020statistical}
Rotem Dror, Lotem Peled-Cohen, Segev Shlomov, and Roi Reichart. 2020.
\newblock Statistical significance testing for natural language processing.
\newblock \emph{Synthesis Lectures on Human Language Technologies},
  13(2):1--116.

\bibitem[{Dror et~al.(2019)Dror, Shlomov, and Reichart}]{dror2019deep}
Rotem Dror, Segev Shlomov, and Roi Reichart. 2019.
\newblock \href {https://doi.org/10.18653/v1/p19-1266} {Deep dominance - how to
  properly compare deep neural models}.
\newblock In \emph{Proceedings of the 57th Conference of the Association for
  Computational Linguistics, {ACL} 2019, Florence, Italy, July 28- August 2,
  2019, Volume 1: Long Papers}, pages 2773--2785. Association for Computational
  Linguistics.

\bibitem[{Drummond(2009)}]{drummond2009repl}
Chris Drummond. 2009.
\newblock Replicability is not reproducibility: Nor is it good science.
\newblock \emph{Proceedings of the Evaluation Methods for Machine Learning
  Workshop at the 26th ICML}.

\bibitem[{ELRA(1995)}]{elra-citation}
ELRA. 1995.
\newblock The european language resources association (elra).
\newblock \url{http://www.elra.info/en/about/}.
\newblock Accessed: 2022-02-10.

\bibitem[{Fokkens et~al.(2013)Fokkens, van Erp, Postma, Pedersen, Vossen, and
  Freire}]{fokkens-etal-2013-offspring}
Antske Fokkens, Marieke van Erp, Marten Postma, Ted Pedersen, Piek Vossen, and
  Nuno Freire. 2013.
\newblock \href {https://aclanthology.org/P13-1166} {Offspring from
  reproduction problems: What replication failure teaches us}.
\newblock In \emph{Proceedings of the 51st Annual Meeting of the Association
  for Computational Linguistics (Volume 1: Long Papers)}, pages 1691--1701,
  Sofia, Bulgaria. Association for Computational Linguistics.

\bibitem[{Gebru et~al.(2020)Gebru, Morgenstern, Vecchione, Vaughan, Wallach,
  III, and Crawford}]{gebru2020}
Timnit Gebru, Jamie Morgenstern, Briana Vecchione, Jennifer~Wortman Vaughan,
  Hanna Wallach, Hal~Daumé III, and Kate Crawford. 2020.
\newblock \href {https://arxiv.org/abs/1803.09010} {Datasheets for datasets}.
\newblock \emph{Computing Research Repository}, arxiv:1803.09010.
\newblock Version 7.

\bibitem[{Gehrmann et~al.(2022)Gehrmann, Clark, and
  Sellam}]{gehrmann2022repairing}
Sebastian Gehrmann, Elizabeth Clark, and Thibault Sellam. 2022.
\newblock Repairing the cracked foundation: A survey of obstacles in evaluation
  practices for generated text.
\newblock \emph{arXiv preprint arXiv:2202.06935}.

\bibitem[{Gelman et~al.(2013)Gelman, Carlin, Stern, Dunson, Vehtari, and
  Rubin}]{gelmanbayesian}
Andrew Gelman, John~B Carlin, Hal~S Stern, David~B Dunson, Aki Vehtari, and
  Donald~B Rubin. 2013.
\newblock Bayesian data analysis.
\newblock \emph{Chapman Hall, London}.

\bibitem[{Gong et~al.(2019)Gong, Zhong, and Hu}]{gong2019}
Zhiqiang Gong, Ping Zhong, and Weidong Hu. 2019.
\newblock Diversity in machine learning.
\newblock \emph{IEEE Access}, 7:64323--64350.

\bibitem[{Gorman and Bedrick(2019)}]{gorman2019we}
Kyle Gorman and Steven Bedrick. 2019.
\newblock We need to talk about standard splits.
\newblock In \emph{Proceedings of the 57th annual meeting of the association
  for computational linguistics}, pages 2786--2791.

\bibitem[{Greenland et~al.(2016)Greenland, Senn, Rothman, Carlin, Poole,
  Goodman, and Altman}]{greenland2016statistical}
Sander Greenland, Stephen~J Senn, Kenneth~J Rothman, John~B Carlin, Charles
  Poole, Steven~N Goodman, and Douglas~G Altman. 2016.
\newblock Statistical tests, p values, confidence intervals, and power: a guide
  to misinterpretations.
\newblock \emph{European journal of epidemiology}, 31(4):337--350.

\bibitem[{Gundersen and Kjensmo(2018)}]{gundersen2018state}
Odd~Erik Gundersen and Sigbj{\o}rn Kjensmo. 2018.
\newblock State of the art: Reproducibility in artificial intelligence.
\newblock In \emph{Proceedings of the AAAI Conference on Artificial
  Intelligence}, volume~32.

\bibitem[{Gururangan et~al.(2020)Gururangan, Marasovi{\'c}, Swayamdipta, Lo,
  Beltagy, Downey, and Smith}]{gururangan-etal-2020-dont}
Suchin Gururangan, Ana Marasovi{\'c}, Swabha Swayamdipta, Kyle Lo, Iz~Beltagy,
  Doug Downey, and Noah~A. Smith. 2020.
\newblock \href {https://doi.org/10.18653/v1/2020.acl-main.740} {Don{'}t stop
  pretraining: Adapt language models to domains and tasks}.
\newblock In \emph{Proceedings of the 58th Annual Meeting of the Association
  for Computational Linguistics}, pages 8342--8360, Online. Association for
  Computational Linguistics.

\bibitem[{Hardmeier et~al.(2021)Hardmeier, Costa{-}juss{\`{a}}, Webster,
  Radford, and Blodgett}]{bias_statement}
Christian Hardmeier, Marta~R. Costa{-}juss{\`{a}}, Kellie Webster, Will
  Radford, and Su~Lin Blodgett. 2021.
\newblock \href {http://arxiv.org/abs/2104.03026} {How to write a bias
  statement: Recommendations for submissions to the workshop on gender bias in
  {NLP}}.
\newblock \emph{CoRR}, abs/2104.03026.

\bibitem[{He et~al.(2019)He, Baxter, Xu, Xu, Zhou, and Zhang}]{he2019practical}
Jianxing He, Sally~L Baxter, Jie Xu, Jiming Xu, Xingtao Zhou, and Kang Zhang.
  2019.
\newblock The practical implementation of artificial intelligence technologies
  in medicine.
\newblock \emph{Nature medicine}, 25(1):30--36.

\bibitem[{Henderson et~al.(2020)Henderson, Hu, Romoff, Brunskill, Jurafsky, and
  Pineau}]{henderson2020towards}
Peter Henderson, Jieru Hu, Joshua Romoff, Emma Brunskill, Dan Jurafsky, and
  Joelle Pineau. 2020.
\newblock Towards the systematic reporting of the energy and carbon footprints
  of machine learning.
\newblock \emph{Journal of Machine Learning Research}, 21(248):1--43.

\bibitem[{Henderson et~al.(2018)Henderson, Islam, Bachman, Pineau, Precup, and
  Meger}]{henderson2018deep}
Peter Henderson, Riashat Islam, Philip Bachman, Joelle Pineau, Doina Precup,
  and David Meger. 2018.
\newblock \href
  {https://www.aaai.org/ocs/index.php/AAAI/AAAI18/paper/view/16669} {Deep
  reinforcement learning that matters}.
\newblock In \emph{Proceedings of the Thirty-Second {AAAI} Conference on
  Artificial Intelligence, (AAAI-18), the 30th innovative Applications of
  Artificial Intelligence (IAAI-18), and the 8th {AAAI} Symposium on
  Educational Advances in Artificial Intelligence (EAAI-18), New Orleans,
  Louisiana, USA, February 2-7, 2018}, pages 3207--3214. {AAAI} Press.

\bibitem[{Hershcovich et~al.(2022)Hershcovich, Webersinke, Kraus, Bingler, and
  Leippold}]{hershcovich2022towards}
Daniel Hershcovich, Nicolas Webersinke, Mathias Kraus, Julia~Anna Bingler, and
  Markus Leippold. 2022.
\newblock Towards climate awareness in nlp research.
\newblock \emph{arXiv preprint arXiv:2205.05071}.

\bibitem[{Hooker(2021)}]{hooker2021hardware}
Sara Hooker. 2021.
\newblock The hardware lottery.
\newblock \emph{Communications of the ACM}, 64(12):58--65.

\bibitem[{Hovy and Prabhumoye(2021)}]{hovyshrimai21}
Dirk Hovy and Shrimai Prabhumoye. 2021.
\newblock Five sources of bias in natural language processing.
\newblock \emph{Language and Linguistics Compass}, 15(8):e12432.

\bibitem[{Hovy and Spruit(2016)}]{hovy-spruit-2016-social}
Dirk Hovy and Shannon~L. Spruit. 2016.
\newblock \href {https://doi.org/10.18653/v1/P16-2096} {The social impact of
  natural language processing}.
\newblock In \emph{Proceedings of the 54th Annual Meeting of the Association
  for Computational Linguistics (Volume 2: Short Papers)}, pages 591--598,
  Berlin, Germany. Association for Computational Linguistics.

\bibitem[{Ilyas et~al.(2019)Ilyas, Santurkar, Tsipras, Engstrom, Tran, and
  Madry}]{ilyas2019adversarial}
Andrew Ilyas, Shibani Santurkar, Dimitris Tsipras, Logan Engstrom, Brandon
  Tran, and Aleksander Madry. 2019.
\newblock \href
  {https://proceedings.neurips.cc/paper/2019/hash/e2c420d928d4bf8ce0ff2ec19b371514-Abstract.html}
  {Adversarial examples are not bugs, they are features}.
\newblock In \emph{Advances in Neural Information Processing Systems 32: Annual
  Conference on Neural Information Processing Systems 2019, NeurIPS 2019,
  December 8-14, 2019, Vancouver, BC, Canada}, pages 125--136.

\bibitem[{Ioannidis(2005)}]{Ioannidis2005}
John P.~A. Ioannidis. 2005.
\newblock \href {https://doi.org/10.1371/journal.pmed.0020124} {Why most
  published research findings are false}.
\newblock \emph{PLOS Medicine}, 2(8):null.

\bibitem[{Japkowicz and Shah(2011)}]{japkowicz2011evaluating}
Nathalie Japkowicz and Mohak Shah. 2011.
\newblock \emph{Evaluating learning algorithms: a classification perspective}.
\newblock Cambridge University Press.

\bibitem[{Jean-Paul et~al.(2019)Jean-Paul, Elseify, Obeid, and
  Picone}]{jean2019issues}
S~Jean-Paul, T~Elseify, I~Obeid, and J~Picone. 2019.
\newblock Issues in the reproducibility of deep learning results.
\newblock In \emph{2019 IEEE Signal Processing in Medicine and Biology
  Symposium (SPMB)}, pages 1--4. IEEE.

\bibitem[{Jensen et~al.(2021)Jensen, Kelly, and Pedersen}]{jensen2021there}
Theis~Ingerslev Jensen, Bryan~T Kelly, and Lasse~Heje Pedersen. 2021.
\newblock Is there a replication crisis in finance?
\newblock Technical report, National Bureau of Economic Research.

\bibitem[{John et~al.(2012)John, Loewenstein, and Prelec}]{john2012measuring}
Leslie~K John, George Loewenstein, and Drazen Prelec. 2012.
\newblock Measuring the prevalence of questionable research practices with
  incentives for truth telling.
\newblock \emph{Psychological science}, 23(5):524--532.

\bibitem[{Joshi et~al.(2020)Joshi, Chen, Liu, Weld, Zettlemoyer, and
  Levy}]{joshi-etal-2020-spanbert}
Mandar Joshi, Danqi Chen, Yinhan Liu, Daniel~S. Weld, Luke Zettlemoyer, and
  Omer Levy. 2020.
\newblock \href {https://doi.org/10.1162/tacl_a_00300} {{S}pan{BERT}: Improving
  pre-training by representing and predicting spans}.
\newblock \emph{Transactions of the Association for Computational Linguistics},
  8:64--77.

\bibitem[{Kiela et~al.(2021)Kiela, Bartolo, Nie, Kaushik, Geiger, Wu, Vidgen,
  Prasad, Singh, Ringshia, Ma, Thrush, Riedel, Waseem, Stenetorp, Jia, Bansal,
  Potts, and Williams}]{kiela-etal-2021-dynabench}
Douwe Kiela, Max Bartolo, Yixin Nie, Divyansh Kaushik, Atticus Geiger,
  Zhengxuan Wu, Bertie Vidgen, Grusha Prasad, Amanpreet Singh, Pratik Ringshia,
  Zhiyi Ma, Tristan Thrush, Sebastian Riedel, Zeerak Waseem, Pontus Stenetorp,
  Robin Jia, Mohit Bansal, Christopher Potts, and Adina Williams. 2021.
\newblock \href {https://doi.org/10.18653/v1/2021.naacl-main.324} {Dynabench:
  Rethinking benchmarking in {NLP}}.
\newblock In \emph{Proceedings of the 2021 Conference of the North American
  Chapter of the Association for Computational Linguistics: Human Language
  Technologies}, pages 4110--4124, Online. Association for Computational
  Linguistics.

\bibitem[{Koh et~al.(2020)Koh, Sagawa, Marklund, Xie, Zhang, Balsubramani, Hu,
  Yasunaga, Phillips, Beery, Leskovec, Kundaje, Pierson, Levine, Finn, and
  Liang}]{wilds2020}
Pang~Wei Koh, Shiori Sagawa, Henrik Marklund, Sang~Michael Xie, Marvin Zhang,
  Akshay Balsubramani, Weihua Hu, Michihiro Yasunaga, Richard~Lanas Phillips,
  Sara Beery, Jure Leskovec, Anshul Kundaje, Emma Pierson, Sergey Levine,
  Chelsea Finn, and Percy Liang. 2020.
\newblock \href {http://arxiv.org/abs/2012.07421} {{WILDS:} {A} benchmark of
  in-the-wild distribution shifts}.
\newblock \emph{CoRR}, abs/2012.07421.

\bibitem[{Kruschke(2010)}]{kruschke2010bayesian}
John~K Kruschke. 2010.
\newblock Bayesian data analysis.
\newblock \emph{Wiley Interdisciplinary Reviews: Cognitive Science},
  1(5):658--676.

\bibitem[{Kruschke(2013)}]{kruschke2013bayesian}
John~K Kruschke. 2013.
\newblock Bayesian estimation supersedes the t test.
\newblock \emph{Journal of Experimental Psychology: General}, 142(2):573.

\bibitem[{Kruschke(2021)}]{kruschke2021bayesian}
John~K Kruschke. 2021.
\newblock Bayesian analysis reporting guidelines.
\newblock \emph{Nature human behaviour}, 5(10):1282--1291.

\bibitem[{Kruschke and Liddell(2018)}]{kruschke2018bayesian}
John~K Kruschke and Torrin~M Liddell. 2018.
\newblock Bayesian data analysis for newcomers.
\newblock \emph{Psychonomic bulletin \& review}, 25(1):155--177.

\bibitem[{Kuhn(1970)}]{kuhn1970structure}
Thomas~S Kuhn. 1970.
\newblock \emph{The structure of scientific revolutions}, volume 111.
\newblock Chicago University of Chicago Press.

\bibitem[{Lampinen et~al.(2021)Lampinen, Chan, Santoro, and
  Hill}]{lampinen2021publishing}
Andrew~Kyle Lampinen, Stephanie~CY Chan, Adam Santoro, and Felix Hill. 2021.
\newblock Publishing fast and slow: A path toward generalizability in
  psychology and ai.

\bibitem[{Leventi-Peetz and {\"O}streich(2022)}]{leventi2022deep}
A-M Leventi-Peetz and T~{\"O}streich. 2022.
\newblock Deep learning reproducibility and explainable ai (xai).
\newblock \emph{arXiv preprint arXiv:2202.11452}.

\bibitem[{Lhoest et~al.(2021)Lhoest, del Moral, von Platen, Wolf, Jernite,
  Thakur, Tunstall, Patil, Drame, Chaumond, Plu, Davison, Brandeis, Scao, Sanh,
  Xu, Patry, McMillan-Major, Schmid, Gugger, Liu, Raw, Lesage, Matussière,
  Debut, Bekman, and Delangue}]{quentin_lhoest_2021_5510481}
Quentin Lhoest, Albert~Villanova del Moral, Patrick von Platen, Thomas Wolf,
  Yacine Jernite, Abhishek Thakur, Lewis Tunstall, Suraj Patil, Mariama Drame,
  Julien Chaumond, Julien Plu, Joe Davison, Simon Brandeis, Teven~Le Scao,
  Victor Sanh, Kevin~Canwen Xu, Nicolas Patry, Angelina McMillan-Major, Philipp
  Schmid, Sylvain Gugger, Steven Liu, Nathan Raw, Sylvain Lesage, Théo
  Matussière, Lysandre Debut, Stas Bekman, and Clément Delangue. 2021.
\newblock \href {https://doi.org/10.5281/zenodo.5510481} {huggingface/datasets:
  1.12.1}.

\bibitem[{Li et~al.(2017)Li, Jamieson, DeSalvo, Rostamizadeh, and
  Talwalkar}]{li2017hyperband}
Lisha Li, Kevin Jamieson, Giulia DeSalvo, Afshin Rostamizadeh, and Ameet
  Talwalkar. 2017.
\newblock Hyperband: A novel bandit-based approach to hyperparameter
  optimization.
\newblock \emph{The Journal of Machine Learning Research}, 18(1):6765--6816.

\bibitem[{Liberman(2015)}]{liberman2015replicability}
Mark Liberman. 2015.
\newblock Replicability vs. reproducibility — or is it the other way around?
\newblock \url{https://languagelog.ldc.upenn.edu/nll/?p=21956}.
\newblock Accessed: 2022-02-21.

\bibitem[{Lin et~al.(2013)Lin, Lucas~Jr, and Shmueli}]{lin2013research}
Mingfeng Lin, Henry~C Lucas~Jr, and Galit Shmueli. 2013.
\newblock Research commentary—too big to fail: large samples and the p-value
  problem.
\newblock \emph{Information Systems Research}, 24(4):906--917.

\bibitem[{Locascio(2017)}]{locascio2017results}
Joseph~J Locascio. 2017.
\newblock Results blind science publishing.
\newblock \emph{Basic and applied social psychology}, 39(5):239--246.

\bibitem[{Manibardo et~al.(2021)Manibardo, La{\~n}a, and
  Del~Ser}]{manibardo2021deep}
Eric~L Manibardo, Ibai La{\~n}a, and Javier Del~Ser. 2021.
\newblock Deep learning for road traffic forecasting: Does it make a
  difference?
\newblock \emph{IEEE Transactions on Intelligent Transportation Systems}.

\bibitem[{Manning(2015)}]{manning-2015-last}
Christopher~D. Manning. 2015.
\newblock \href {https://doi.org/doi:10.1162/COLI_a_00239} {Last words:
  Computational linguistics and deep learning}.
\newblock \emph{Computational Linguistics}, 41(4):701--707.

\bibitem[{Marie et~al.(2021)Marie, Fujita, and Rubino}]{marie2021scientific}
Benjamin Marie, Atsushi Fujita, and Raphael Rubino. 2021.
\newblock \href {https://doi.org/10.18653/v1/2021.acl-long.566} {Scientific
  credibility of machine translation research: {A} meta-evaluation of 769
  papers}.
\newblock In \emph{Proceedings of the 59th Annual Meeting of the Association
  for Computational Linguistics and the 11th International Joint Conference on
  Natural Language Processing, {ACL/IJCNLP} 2021, (Volume 1: Long Papers),
  Virtual Event, August 1-6, 2021}, pages 7297--7306. Association for
  Computational Linguistics.

\bibitem[{Menon et~al.(2020)Menon, Damian, Hu, Ravi, and
  Rudin}]{menon2020pulse}
Sachit Menon, Alexandru Damian, Shijia Hu, Nikhil Ravi, and Cynthia Rudin.
  2020.
\newblock Pulse: Self-supervised photo upsampling via latent space exploration
  of generative models.
\newblock In \emph{Proceedings of the ieee/cvf conference on computer vision
  and pattern recognition}, pages 2437--2445.

\bibitem[{Mitchell et~al.(2019)Mitchell, Wu, Zaldivar, Barnes, Vasserman,
  Hutchinson, Spitzer, Raji, and Gebru}]{mitchell2019model}
Margaret Mitchell, Simone Wu, Andrew Zaldivar, Parker Barnes, Lucy Vasserman,
  Ben Hutchinson, Elena Spitzer, Inioluwa~Deborah Raji, and Timnit Gebru. 2019.
\newblock Model cards for model reporting.
\newblock In \emph{Proceedings of the conference on fairness, accountability,
  and transparency}, pages 220--229.

\bibitem[{Mohamed et~al.(2020)Mohamed, Png, and Isaac}]{mohamed2020decolonial}
Shakir Mohamed, Marie-Therese Png, and William Isaac. 2020.
\newblock Decolonial ai: Decolonial theory as sociotechnical foresight in
  artificial intelligence.
\newblock \emph{Philosophy \& Technology}, 33(4):659--684.

\bibitem[{Mohammad(2020)}]{panelMohammad}
Saif~M. Mohammad. 2020.
\newblock \href {http://www.saifmohammad.com/WebDocs/EthicsStatement-web.pdf}
  {What is a research ethics statement and why does it matter?}

\bibitem[{Mohammad(2021)}]{mohammad2021ethics}
Saif~M. Mohammad. 2021.
\newblock \href {http://arxiv.org/abs/2107.01183} {Ethics sheets for {AI}
  tasks}.
\newblock \emph{CoRR}, abs/2107.01183.

\bibitem[{Nagarajan et~al.(2021)Nagarajan, Andreassen, and
  Neyshabur}]{nagarajan2021understanding}
Vaishnavh Nagarajan, Anders Andreassen, and Behnam Neyshabur. 2021.
\newblock \href {https://openreview.net/forum?id=fSTD6NFIW\_b} {Understanding
  the failure modes of out-of-distribution generalization}.
\newblock In \emph{9th International Conference on Learning Representations,
  {ICLR} 2021, Virtual Event, Austria, May 3-7, 2021}. OpenReview.net.

\bibitem[{Narang et~al.(2021)Narang, Chung, Tay, Fedus, F{\'{e}}vry, Matena,
  Malkan, Fiedel, Shazeer, Lan, Zhou, Li, Ding, Marcus, Roberts, and
  Raffel}]{narang2021transformer}
Sharan Narang, Hyung~Won Chung, Yi~Tay, Liam Fedus, Thibault F{\'{e}}vry,
  Michael Matena, Karishma Malkan, Noah Fiedel, Noam Shazeer, Zhenzhong Lan,
  Yanqi Zhou, Wei Li, Nan Ding, Jake Marcus, Adam Roberts, and Colin Raffel.
  2021.
\newblock \href {https://doi.org/10.18653/v1/2021.emnlp-main.465} {Do
  transformer modifications transfer across implementations and applications?}
\newblock In \emph{Proceedings of the 2021 Conference on Empirical Methods in
  Natural Language Processing, {EMNLP} 2021, Virtual Event / Punta Cana,
  Dominican Republic, 7-11 November, 2021}, pages 5758--5773. Association for
  Computational Linguistics.

\bibitem[{Nilsson et~al.(2018)Nilsson, Smith, Ulm, Gustavsson, and
  Jirstrand}]{nilsson2018performance}
Adrian Nilsson, Simon Smith, Gregor Ulm, Emil Gustavsson, and Mats Jirstrand.
  2018.
\newblock A performance evaluation of federated learning algorithms.
\newblock In \emph{Proceedings of the second workshop on distributed
  infrastructures for deep learning}, pages 1--8.

\bibitem[{Nivre et~al.(2020)Nivre, de~Marneffe, Ginter, Haji{\v{c}}, Manning,
  Pyysalo, Schuster, Tyers, and Zeman}]{nivre-etal-2020-universal}
Joakim Nivre, Marie-Catherine de~Marneffe, Filip Ginter, Jan Haji{\v{c}},
  Christopher~D. Manning, Sampo Pyysalo, Sebastian Schuster, Francis Tyers, and
  Daniel Zeman. 2020.
\newblock \href {https://aclanthology.org/2020.lrec-1.497} {{U}niversal
  {D}ependencies v2: An evergrowing multilingual treebank collection}.
\newblock In \emph{Proceedings of the 12th Language Resources and Evaluation
  Conference}, pages 4034--4043, Marseille, France. European Language Resources
  Association.

\bibitem[{Nosek et~al.(2018)Nosek, Ebersole, DeHaven, and
  Mellor}]{nosek2018preregistration}
Brian~A Nosek, Charles~R Ebersole, Alexander~C DeHaven, and David~T Mellor.
  2018.
\newblock The preregistration revolution.
\newblock \emph{Proceedings of the National Academy of Sciences},
  115(11):2600--2606.

\bibitem[{Parmar et~al.(2022)Parmar, Mishra, Geva, and Baral}]{parmar2022don}
Mihir Parmar, Swaroop Mishra, Mor Geva, and Chitta Baral. 2022.
\newblock Don't blame the annotator: Bias already starts in the annotation
  instructions.
\newblock \emph{arXiv preprint arXiv:2205.00415}.

\bibitem[{Paun et~al.(2022)Paun, Artstein, and Poesio}]{paun2022statistical}
Silviu Paun, Ron Artstein, and Massimo Poesio. 2022.
\newblock Statistical methods for annotation analysis.
\newblock \emph{Synthesis Lectures on Human Language Technologies},
  15(1):1--217.

\bibitem[{Pedersen(2008)}]{pedersen-2008-last}
Ted Pedersen. 2008.
\newblock \href {https://doi.org/10.1162/coli.2008.34.3.465} {Last words:
  Empiricism is not a matter of faith}.
\newblock \emph{Computational Linguistics}, 34(3):465--470.

\bibitem[{Peng(2011)}]{peng2011reproducible}
Roger~D Peng. 2011.
\newblock Reproducible research in computational science.
\newblock \emph{Science}, 334(6060):1226--1227.

\bibitem[{Pfeiffer et~al.(2020)Pfeiffer, R{\"u}ckl{\'e}, Poth, Kamath,
  Vuli{\'c}, Ruder, Cho, and Gurevych}]{pfeiffer2020AdapterHub}
Jonas Pfeiffer, Andreas R{\"u}ckl{\'e}, Clifton Poth, Aishwarya Kamath, Ivan
  Vuli{\'c}, Sebastian Ruder, Kyunghyun Cho, and Iryna Gurevych. 2020.
\newblock Adapterhub: A framework for adapting transformers.
\newblock In \emph{Proceedings of the 2020 Conference on Empirical Methods in
  Natural Language Processing: System Demonstrations}, pages 46--54.

\bibitem[{Plank et~al.(2020)Plank, Jensen, and van~der
  Goot}]{plank-etal-2020-dan+}
Barbara Plank, Kristian~N{\o}rgaard Jensen, and Rob van~der Goot. 2020.
\newblock {D}a{N}+: {D}anish nested named entities and lexical normalization.
\newblock In \emph{Proceedings of the 28th International Conference on
  Computational Linguistics}, pages 6649--6662, Barcelona, Spain (Online).
  International Committee on Computational Linguistics.

\bibitem[{Popper(1934)}]{popper1934}
Karl Popper. 1934.
\newblock \emph{Karl Popper: Logik der Forschung}.
\newblock Mohr Siebeck, T\"ubingen, Germany.

\bibitem[{Post(2018)}]{post-2018-call}
Matt Post. 2018.
\newblock \href {https://doi.org/10.18653/v1/W18-6319} {A call for clarity in
  reporting {BLEU} scores}.
\newblock In \emph{Proceedings of the Third Conference on Machine Translation:
  Research Papers}, pages 186--191, Brussels, Belgium. Association for
  Computational Linguistics.

\bibitem[{Prabhakaran et~al.(2021)Prabhakaran, Mostafazadeh~Davani, and
  Diaz}]{prabhakaran-etal-2021-releasing}
Vinodkumar Prabhakaran, Aida Mostafazadeh~Davani, and Mark Diaz. 2021.
\newblock \href {https://doi.org/10.18653/v1/2021.law-1.14} {On releasing
  annotator-level labels and information in datasets}.
\newblock In \emph{Proceedings of The Joint 15th Linguistic Annotation Workshop
  (LAW) and 3rd Designing Meaning Representations (DMR) Workshop}, pages
  133--138, Punta Cana, Dominican Republic. Association for Computational
  Linguistics.

\bibitem[{Pustejovsky and Stubbs(2012)}]{pustejovsky2012natural}
James Pustejovsky and Amber Stubbs. 2012.
\newblock \emph{Natural Language Annotation for Machine Learning: A guide to
  corpus-building for applications}.
\newblock O'Reilly Media, Inc.

\bibitem[{Raff(2022)}]{raff2022does}
Edward Raff. 2022.
\newblock Does the market of citations reward reproducible work?
\newblock \emph{arXiv preprint arXiv:2204.03829}.

\bibitem[{Ramesh~Kashyap et~al.(2021)Ramesh~Kashyap, Hazarika, Kan, and
  Zimmermann}]{rameshkashyap2021}
Abhinav Ramesh~Kashyap, Devamanyu Hazarika, Min-Yen Kan, and Roger Zimmermann.
  2021.
\newblock \href {https://doi.org/10.18653/v1/2021.naacl-main.147} {Domain
  divergences: A survey and empirical analysis}.
\newblock In \emph{Proceedings of the 2021 Conference of the North American
  Chapter of the Association for Computational Linguistics: Human Language
  Technologies}, pages 1830--1849, Online. Association for Computational
  Linguistics.

\bibitem[{Raschka(2018)}]{raschka2018model}
Sebastian Raschka. 2018.
\newblock Model evaluation, model selection, and algorithm selection in machine
  learning.
\newblock \emph{arXiv preprint arXiv:1811.12808}.

\bibitem[{Rei(2022)}]{rei2022publications}
Marek Rei. 2022.
\newblock Ml and nlp publications in 2021.
\newblock \url{https://www.marekrei.com/blog/ml-and-nlp-publications-in-2021/}.

\bibitem[{Reitz and Schlusser(2016)}]{reitz2016hitchhiker}
Kenneth Reitz and Tanya Schlusser. 2016.
\newblock \emph{The Hitchhiker's guide to Python: best practices for
  development}.
\newblock " O'Reilly Media, Inc.".

\bibitem[{Ribeiro et~al.(2020)Ribeiro, Wu, Guestrin, and
  Singh}]{ribeiro-etal-2020-beyond}
Marco~Tulio Ribeiro, Tongshuang Wu, Carlos Guestrin, and Sameer Singh. 2020.
\newblock \href {https://doi.org/10.18653/v1/2020.acl-main.442} {Beyond
  accuracy: Behavioral testing of {NLP} models with {C}heck{L}ist}.
\newblock In \emph{Proceedings of the 58th Annual Meeting of the Association
  for Computational Linguistics}, pages 4902--4912, Online. Association for
  Computational Linguistics.

\bibitem[{Riezler and Hagmann(2021)}]{riezler2021validity}
Stefan Riezler and Michael Hagmann. 2021.
\newblock Validity, reliability, and significance.

\bibitem[{Riezler and Maxwell~III(2005)}]{riezler2005some}
Stefan Riezler and John~T Maxwell~III. 2005.
\newblock On some pitfalls in automatic evaluation and significance testing for
  mt.
\newblock In \emph{Proceedings of the ACL workshop on intrinsic and extrinsic
  evaluation measures for machine translation and/or summarization}, pages
  57--64.

\bibitem[{Rogers and Augenstein(2020)}]{rogers2020what}
Anna Rogers and Isabelle Augenstein. 2020.
\newblock \href {https://doi.org/10.18653/v1/2020.findings-emnlp.112} {What can
  we do to improve peer review in nlp?}
\newblock In \emph{Findings of the Association for Computational Linguistics:
  {EMNLP} 2020, Online Event, 16-20 November 2020}, volume {EMNLP} 2020 of
  \emph{Findings of {ACL}}, pages 1256--1262. Association for Computational
  Linguistics.

\bibitem[{Rogers and Augenstein(2021)}]{rogers2021how}
Anna Rogers and Isabelle Augenstein. 2021.
\newblock How to review for acl rolling review?
\newblock \url{https://aclrollingreview.org/reviewertutorial}.
\newblock Accessed: 2022-02-21.

\bibitem[{Sap et~al.(2021)Sap, Swayamdipta, Vianna, Zhou, Choi, and
  Smith}]{sap2021annotators}
Maarten Sap, Swabha Swayamdipta, Laura Vianna, Xuhui Zhou, Yejin Choi, and
  Noah~A Smith. 2021.
\newblock Annotators with attitudes: How annotator beliefs and identities bias
  toxic language detection.
\newblock \emph{arXiv preprint arXiv:2111.07997}.

\bibitem[{Schlangen(2021)}]{schlangen-2021-targeting}
David Schlangen. 2021.
\newblock \href {https://doi.org/10.18653/v1/2021.acl-short.85} {Targeting the
  benchmark: On methodology in current natural language processing research}.
\newblock In \emph{Proceedings of the 59th Annual Meeting of the Association
  for Computational Linguistics and the 11th International Joint Conference on
  Natural Language Processing (Volume 2: Short Papers)}, pages 670--674,
  Online. Association for Computational Linguistics.

\bibitem[{Schmidt et~al.(2021)Schmidt, Goyal, Joshi, Feld, Conell, Laskaris,
  Blank, Wilson, Friedler, and Luccioni}]{codecarbon}
Victor Schmidt, Kamal Goyal, Aditya Joshi, Boris Feld, Liam Conell, Nikolas
  Laskaris, Doug Blank, Jonathan Wilson, Sorelle Friedler, and Sasha Luccioni.
  2021.
\newblock \href {https://doi.org/10.5281/zenodo.4658424} {{CodeCarbon: Estimate
  and Track Carbon Emissions from Machine Learning Computing}}.

\bibitem[{Schwartz et~al.(2020)Schwartz, Dodge, Smith, and
  Etzioni}]{schwartz2020green}
Roy Schwartz, Jesse Dodge, Noah~A. Smith, and Oren Etzioni. 2020.
\newblock \href {https://doi.org/10.1145/3381831} {Green {AI}}.
\newblock \emph{Commun. {ACM}}, 63(12):54--63.

\bibitem[{Sellam et~al.(2021)Sellam, Yadlowsky, Wei, Saphra, D'Amour, Linzen,
  Bastings, Turc, Eisenstein, Das et~al.}]{sellam2021multiberts}
Thibault Sellam, Steve Yadlowsky, Jason Wei, Naomi Saphra, Alexander D'Amour,
  Tal Linzen, Jasmijn Bastings, Iulia Turc, Jacob Eisenstein, Dipanjan Das,
  et~al. 2021.
\newblock The multiberts: Bert reproductions for robustness analysis.
\newblock \emph{arXiv preprint arXiv:2106.16163}.

\bibitem[{Sevilla et~al.(2022)Sevilla, Heim, Ho, Besiroglu, Hobbhahn, and
  Villalobos}]{sevilla2022compute}
Jaime Sevilla, Lennart Heim, Anson Ho, Tamay Besiroglu, Marius Hobbhahn, and
  Pablo Villalobos. 2022.
\newblock Compute trends across three eras of machine learning.
\newblock \emph{arXiv:2202.05924 [cs]}.
\newblock ArXiv: 2202.05924.

\bibitem[{Shimorina et~al.(2021)Shimorina, Parmentier, and
  Gardent}]{shimorina2021error}
Anastasia Shimorina, Yannick Parmentier, and Claire Gardent. 2021.
\newblock An error analysis framework for shallow surface realization.
\newblock \emph{Transactions of the Association for Computational Linguistics},
  9:429--446.

\bibitem[{Simon(1995)}]{simon1995artificial}
Herbert~A Simon. 1995.
\newblock Artificial intelligence: an empirical science.
\newblock \emph{Artificial Intelligence}, 77(1):95--127.

\bibitem[{Snoek et~al.(2012)Snoek, Larochelle, and Adams}]{snoek2012practical}
Jasper Snoek, Hugo Larochelle, and Ryan~P. Adams. 2012.
\newblock \href
  {https://proceedings.neurips.cc/paper/2012/hash/05311655a15b75fab86956663e1819cd-Abstract.html}
  {Practical bayesian optimization of machine learning algorithms}.
\newblock In \emph{Advances in Neural Information Processing Systems 25: 26th
  Annual Conference on Neural Information Processing Systems 2012. Proceedings
  of a meeting held December 3-6, 2012, Lake Tahoe, Nevada, United States},
  pages 2960--2968.

\bibitem[{S{\o}gaard et~al.(2021)S{\o}gaard, Ebert, Bastings, and
  Filippova}]{sogaard2021we}
Anders S{\o}gaard, Sebastian Ebert, Jasmijn Bastings, and Katja Filippova.
  2021.
\newblock \href {https://www.aclweb.org/anthology/2021.eacl-main.156/} {We need
  to talk about random splits}.
\newblock In \emph{Proceedings of the 16th Conference of the European Chapter
  of the Association for Computational Linguistics: Main Volume, {EACL} 2021,
  Online, April 19 - 23, 2021}, pages 1823--1832. Association for Computational
  Linguistics.

\bibitem[{Specia(2021)}]{lucia-keynote}
Lucia Specia. 2021.
\newblock \href {https://nodalida2021.github.io/invited_speakers.html}
  {Disagreement in human evaluation: blame the task not the annotators.}
\newblock NoDaLiDa keynote.

\bibitem[{Strubell et~al.(2019)Strubell, Ganesh, and
  McCallum}]{strubell-etal-2019-energy}
Emma Strubell, Ananya Ganesh, and Andrew McCallum. 2019.
\newblock \href {https://doi.org/10.18653/v1/P19-1355} {Energy and policy
  considerations for deep learning in {NLP}}.
\newblock In \emph{Proceedings of the 57th Annual Meeting of the Association
  for Computational Linguistics}, pages 3645--3650, Florence, Italy.
  Association for Computational Linguistics.

\bibitem[{Su(2021)}]{su2021you}
Weijie Su. 2021.
\newblock You are the best reviewer of your own papers: An owner-assisted
  scoring mechanism.
\newblock \emph{Advances in Neural Information Processing Systems}, 34.

\bibitem[{TMLR(2022)}]{tmlr2022guidelines}
TMLR. 2022.
\newblock Submission guidelines and editorial policies.
\newblock \url{https://jmlr.org/tmlr/editorial-policies.html}.
\newblock Accessed: 2022-02-09.

\bibitem[{Tseng et~al.(2020)Tseng, Stent, and Maida}]{bestpractices}
Tina Tseng, Amanda Stent, and Domenic Maida. 2020.
\newblock \href {http://arxiv.org/abs/2009.11654} {Best practices for managing
  data annotation projects}.
\newblock \emph{CoRR}, abs/2009.11654.

\bibitem[{Tuff{\'e}ry(2011)}]{tuffery2011data}
St{\'e}phane Tuff{\'e}ry. 2011.
\newblock \emph{Data mining and statistics for decision making}.
\newblock John Wiley \& Sons.

\bibitem[{Turner(2020)}]{turner2020gpt3}
Elliot Turner. 2020.
\newblock Twitter post (@eturner303): Reading the openai gpt-3 paper.
\newblock \url{https://twitter.com/eturner303/status/1266264358771757057}.
\newblock Accessed: 2022-02-09.

\bibitem[{Ulmer et~al.(2022)Ulmer, Hardmeier, and Frellsen}]{ulmer2022deep}
Dennis Ulmer, Christian Hardmeier, and Jes Frellsen. 2022.
\newblock deep-significance-easy and meaningful statistical significance
  testing in the age of neural networks.
\newblock \emph{arXiv preprint arXiv:2204.06815}.

\bibitem[{Uma et~al.(2021)Uma, Fornaciari, Hovy, Paun, Plank, and
  Poesio}]{uma2021learning}
Alexandra~N Uma, Tommaso Fornaciari, Dirk Hovy, Silviu Paun, Barbara Plank, and
  Massimo Poesio. 2021.
\newblock Learning from disagreement: A survey.
\newblock \emph{Journal of Artificial Intelligence Research}, 72:1385--1470.

\bibitem[{van~der Goot(2021)}]{groot2021we}
Rob van~der Goot. 2021.
\newblock \href {https://aclanthology.org/2021.emnlp-main.368} {We need to talk
  about train-dev-test splits}.
\newblock In \emph{Proceedings of the 2021 Conference on Empirical Methods in
  Natural Language Processing, {EMNLP} 2021, Virtual Event / Punta Cana,
  Dominican Republic, 7-11 November, 2021}, pages 4485--4494. Association for
  Computational Linguistics.

\bibitem[{van~der Goot et~al.(2021)van~der Goot, {\"U}st{\"u}n, Ramponi,
  Sharaf, and Plank}]{van-der-goot-etal-2021-massive}
Rob van~der Goot, Ahmet {\"U}st{\"u}n, Alan Ramponi, Ibrahim Sharaf, and
  Barbara Plank. 2021.
\newblock \href {https://doi.org/10.18653/v1/2021.eacl-demos.22} {Massive
  choice, ample tasks ({M}a{C}h{A}mp): A toolkit for multi-task learning in
  {NLP}}.
\newblock In \emph{Proceedings of the 16th Conference of the European Chapter
  of the Association for Computational Linguistics: System Demonstrations},
  pages 176--197, Online. Association for Computational Linguistics.

\bibitem[{van Miltenburg et~al.(2021)van Miltenburg, van~der Lee, and
  Krahmer}]{miltenburg2021preregistering}
Emiel van Miltenburg, Chris van~der Lee, and Emiel Krahmer. 2021.
\newblock \href {https://doi.org/10.18653/v1/2021.naacl-main.51}
  {Preregistering {NLP} research}.
\newblock In \emph{Proceedings of the 2021 Conference of the North American
  Chapter of the Association for Computational Linguistics: Human Language
  Technologies, {NAACL-HLT} 2021, Online, June 6-11, 2021}, pages 613--623.
  Association for Computational Linguistics.

\bibitem[{Varab and Schluter(2020)}]{varab-schluter-2020-da}
Daniel Varab and Natalie Schluter. 2020.
\newblock {D}a{N}ewsroom: A large-scale {D}anish summarisation dataset.
\newblock In \emph{Proceedings of the 12th Language Resources and Evaluation
  Conference}, pages 6731--6739, Marseille, France. European Language Resources
  Association.

\bibitem[{Varab and Schluter(2021)}]{varab-schluter-2021-massivesumm}
Daniel Varab and Natalie Schluter. 2021.
\newblock \href {https://doi.org/10.18653/v1/2021.emnlp-main.797}
  {{M}assive{S}umm: a very large-scale, very multilingual, news summarisation
  dataset}.
\newblock In \emph{Proceedings of the 2021 Conference on Empirical Methods in
  Natural Language Processing}, pages 10150--10161, Online and Punta Cana,
  Dominican Republic. Association for Computational Linguistics.

\bibitem[{V{\'a}radi et~al.(2008)V{\'a}radi, Wittenburg, Krauwer, Wynne, and
  Koskenniemi}]{varadi2008clarin}
Tam{\'a}s V{\'a}radi, Peter Wittenburg, Steven Krauwer, Martin Wynne, and Kimmo
  Koskenniemi. 2008.
\newblock Clarin: Common language resources and technology infrastructure.
\newblock In \emph{6th International Conference on Language Resources and
  Evaluation (LREC 2008)}.

\bibitem[{Vaswani et~al.(2017)Vaswani, Shazeer, Parmar, Uszkoreit, Jones,
  Gomez, Kaiser, and Polosukhin}]{vaswani2017attention}
Ashish Vaswani, Noam Shazeer, Niki Parmar, Jakob Uszkoreit, Llion Jones,
  Aidan~N Gomez, {\L}ukasz Kaiser, and Illia Polosukhin. 2017.
\newblock Attention is all you need.
\newblock In \emph{Advances in neural information processing systems}, pages
  5998--6008.

\bibitem[{Waseem et~al.(2021)Waseem, Lulz, Bingel, and
  Augenstein}]{waseem2021disembodied}
Zeerak Waseem, Smarika Lulz, Joachim Bingel, and Isabelle Augenstein. 2021.
\newblock \href {http://arxiv.org/abs/2101.11974} {Disembodied machine
  learning: On the illusion of objectivity in nlp}.

\bibitem[{Wasserstein et~al.(2019)Wasserstein, Schirm, and
  Lazar}]{wasserstein2019moving}
Ronald~L Wasserstein, Allen~L Schirm, and Nicole~A Lazar. 2019.
\newblock Moving to a world beyond “p \textless 0.05”.

\bibitem[{Wei et~al.(2020)Wei, Zhang, Zhou, Li, and Al~Faruque}]{wei2020leaky}
Junyi Wei, Yicheng Zhang, Zhe Zhou, Zhou Li, and Mohammad~Abdullah Al~Faruque.
  2020.
\newblock Leaky dnn: Stealing deep-learning model secret with gpu
  context-switching side-channel.
\newblock In \emph{2020 50th Annual IEEE/IFIP International Conference on
  Dependable Systems and Networks (DSN)}, pages 125--137. IEEE.

\bibitem[{White and Cotterell(2021)}]{white2021examining}
Jennifer~C. White and Ryan Cotterell. 2021.
\newblock \href {https://doi.org/10.18653/v1/2021.acl-long.38} {Examining the
  inductive bias of neural language models with artificial languages}.
\newblock In \emph{Proceedings of the 59th Annual Meeting of the Association
  for Computational Linguistics and the 11th International Joint Conference on
  Natural Language Processing, {ACL/IJCNLP} 2021, (Volume 1: Long Papers),
  Virtual Event, August 1-6, 2021}, pages 454--463. Association for
  Computational Linguistics.

\bibitem[{Wieling et~al.(2018)Wieling, Rawee, and van
  Noord}]{wieling2018reproducibility}
Martijn Wieling, Josine Rawee, and Gertjan van Noord. 2018.
\newblock Reproducibility in computational linguistics: are we willing to
  share?
\newblock \emph{Computational Linguistics}, 44(4):641--649.

\bibitem[{Wilcoxon(1992)}]{wilcoxon1992individual}
Frank Wilcoxon. 1992.
\newblock Individual comparisons by ranking methods.
\newblock In \emph{Breakthroughs in statistics}, pages 196--202. Springer.

\bibitem[{Wolf et~al.(2020)Wolf, Debut, Sanh, Chaumond, Delangue, Moi, Cistac,
  Rault, Louf, Funtowicz, Davison, Shleifer, von Platen, Ma, Jernite, Plu, Xu,
  Le~Scao, Gugger, Drame, Lhoest, and Rush}]{wolf-etal-2020-transformers}
Thomas Wolf, Lysandre Debut, Victor Sanh, Julien Chaumond, Clement Delangue,
  Anthony Moi, Pierric Cistac, Tim Rault, Remi Louf, Morgan Funtowicz, Joe
  Davison, Sam Shleifer, Patrick von Platen, Clara Ma, Yacine Jernite, Julien
  Plu, Canwen Xu, Teven Le~Scao, Sylvain Gugger, Mariama Drame, Quentin Lhoest,
  and Alexander Rush. 2020.
\newblock \href {https://doi.org/10.18653/v1/2020.emnlp-demos.6} {Transformers:
  State-of-the-art natural language processing}.
\newblock In \emph{Proceedings of the 2020 Conference on Empirical Methods in
  Natural Language Processing: System Demonstrations}, pages 38--45, Online.
  Association for Computational Linguistics.

\bibitem[{Yang et~al.(2018)Yang, Liang, and Zhang}]{yang-etal-2018-design}
Jie Yang, Shuailong Liang, and Yue Zhang. 2018.
\newblock \href {https://aclanthology.org/C18-1327} {Design challenges and
  misconceptions in neural sequence labeling}.
\newblock In \emph{Proceedings of the 27th International Conference on
  Computational Linguistics}, pages 3879--3889, Santa Fe, New Mexico, USA.
  Association for Computational Linguistics.

\bibitem[{Ye et~al.(2021)Ye, Li, Hong, Bai, Chen, Zhou, and Li}]{ye2021ood}
Nanyang Ye, Kaican Li, Lanqing Hong, Haoyue Bai, Yiting Chen, Fengwei Zhou, and
  Zhenguo Li. 2021.
\newblock Ood-bench: Benchmarking and understanding out-of-distribution
  generalization datasets and algorithms.
\newblock \emph{arXiv preprint arXiv:2106.03721}.

\bibitem[{Yu et~al.(2020)Yu, Chen, Du, Li, Rashwan, Hou, Jin, Yang, Liu, Kim,
  and Li}]{tensorflowmodelgarden2020}
Hongkun Yu, Chen Chen, Xianzhi Du, Yeqing Li, Abdullah Rashwan, Le~Hou,
  Pengchong Jin, Fan Yang, Frederick Liu, Jaeyoun Kim, and Jing Li. 2020.
\newblock {TensorFlow Model Garden}.
\newblock \url{https://github.com/tensorflow/models}.

\bibitem[{Yuan and Hayashi(2003)}]{yuan2003bootstrap}
Ke-Hai Yuan and Kentaro Hayashi. 2003.
\newblock Bootstrap approach to inference and power analysis based on three
  test statistics for covariance structure models.
\newblock \emph{British Journal of Mathematical and Statistical Psychology},
  56(1):93--110.

\bibitem[{Zhang and Plank(2021)}]{zhang2021cartography}
Mike Zhang and Barbara Plank. 2021.
\newblock \href {https://doi.org/10.18653/v1/2021.findings-emnlp.36}
  {Cartography active learning}.
\newblock In \emph{Findings of the Association for Computational Linguistics:
  {EMNLP} 2021, Virtual Event / Punta Cana, Dominican Republic, 16-20 November,
  2021}, pages 395--406. Association for Computational Linguistics.

\bibitem[{Zhou et~al.(2015)Zhou, Tian, Sukhbaatar, Szlam, and
  Fergus}]{zhou2015simple}
Bolei Zhou, Yuandong Tian, Sainbayar Sukhbaatar, Arthur Szlam, and Rob Fergus.
  2015.
\newblock Simple baseline for visual question answering.
\newblock \emph{arXiv preprint arXiv:1512.02167}.

\bibitem[{Ziliak and McCloskey(2008)}]{ziliak2008cult}
Steve Ziliak and Deirdre~Nansen McCloskey. 2008.
\newblock \emph{The cult of statistical significance: How the standard error
  costs us jobs, justice, and lives}.
\newblock University of Michigan Press.

\bibitem[{Zong et~al.(2020)Zong, Baheti, Xu, and Ritter}]{zong2020}
Shi Zong, Ashutosh Baheti, Wei Xu, and Alan Ritter. 2020.
\newblock \href {http://arxiv.org/abs/2006.02567} {Extracting {COVID-19} events
  from twitter}.
\newblock \emph{CoRR}, abs/2006.02567.

\end{thebibliography}
\bibliographystyle{acl_natbib}

\appendix
\section{Case Studies \& Further Reading}\label{app:case-studies}

The implementation of the methods we advocate for in our work can be challenging. This is why we dedicate this appendix to listing further resources and pointing to examples that illustrate their intended use.

\subsection{Data}\label{app:case-studies-data}

\paragraph{Data Statement} Following~\citet{bender2018data}, the long form data statement should outline \textsc{Curation Rationale}, \textsc{Language Variety}, \textsc{Speaker Demographic}, \textsc{Annotator Demographic}, \textsc{Speech Situation}, \textsc{Text Characteristics} and a \textsc{Provenance Appendix}. A good example of a long form data statement can be found in Appendix B in~\citet{plank-etal-2020-dan+}, where each of the former mentioned topics are outlined. For example, with respect to \textsc{Annotator Demographic}, they mention ``three students and one faculty (age range: 25-40), gender: male and female. White European. Native language: Danish, German. Socioeconomic status: higher-education student and university faculty.'' This is a concise explanation of the annotators involved in their project.

\paragraph{Data Quality} Text corpora today are building blocks for many downstream NLP applications like question answering and summarization. In the work of~\citet{caswell2021quality}, they audit the quality of quality of 205 language-specific corpora released within major public datasets. At least 15 of these 205 corpora have no usable text, and a large fraction contains less than 50\% sentences of acceptable quality. The tacit recommendation is looking at samples of any dataset before using it or releasing it to the public. A good example is~\citet{varab-schluter-2020-da,varab-schluter-2021-massivesumm}, who filter out low-quality news articles from their summarization dataset with empty summaries or bodies, removing duplicates, and removing summaries that are long than them main body of text. More wide varieties of data filtering can be applied, like filtering on length-ratio, LangID, and TF-IDF wordlists~\cite{caswell-etal-2020-language}. Note that there is no easy solution---data cleaning is not a trivial task~\cite{caswell2021quality}.

\paragraph{Universal Dependencies} \citet{nivre-etal-2020-universal} aims to annotate syntactic dependencies in addition to part-of-speech tags, morphological features etc.\ for as many languages as possible within a consistent set of guidelines. The dataset which consists of treebanks contributed by various authors is updated in a regular half-yearly cycle and is hosted on the long-term storage LINDAT/CLARIN repository \citep{varadi2008clarin}. Each release is clearly versioned such that fair comparisons can be made even while guidelines are continuously adapted. Maintenance of the project is conducted on a public \texttt{git} repository, such that changes to both the data and the guidelines can be followed transparently. This allows for contributors to suggest changes via pull requests.

\subsection{Models}
% \todo{Good examples of Models, Model evaluation, Model cards}
There are several libraries that allow for model hosting or distribution of model weights for ``mature'' models. 
\texttt{HuggingFace}~\cite{wolf-etal-2020-transformers} is an example of hosting models for distribution. It is an easy-to-use library for practitioners in the field. Other examples of model distribution is \texttt{Keras} \texttt{Applications}\footnote{\url{https://keras.io/api/applications/}} or \texttt{TensorFlow Model} \texttt{Garden}~\cite{tensorflowmodelgarden2020}. Other ways of distributing models is setting hyperlinks in the repository (e.g.,~\citealp{joshi-etal-2020-spanbert}), to load the models from the checkpoints they have been saved to. A common denominator of all the aforementioned libraries is to list relevant model performances (designated metrics per task), the model size (in bytes), model parameters (e.g., in millions), and inference time (e.g., any time variable). 

\subsection{Codebase}
At the code-level, there are several examples of codebases with strong documentation and clean project structure. We define documentation and project structure in~\cref{app:readme}. Here, we give examples going from smaller projects to larger Python projects:

The codebase of CateNETS~\cite{curth2021inductive,curth2021nonparametric,curth2021really}\footnote{\url{https://github.com/AliciaCurth/CATENets}} shows a clear project structure. This includes unit tests, versioning of the library, and licensing. In addition, there are specific files for each published work to replicate the results.

Not all projects require a \texttt{pip} installation or unit tests. For example---similar to the previous project---MaChAmp~\cite{van-der-goot-etal-2021-massive}\footnote{\url{https://github.com/machamp-nlp/machamp}} shows detailed documentation, including several reproducible experiments shown in the paper (including files with model scores) and a clear project structure. Here, one possible complication lies in possible dependency issues once the repository grows, with unit tests as a mitigation strategy.
%\todo{wow, wish we had the time for that ;)}

AdapterHub~\cite{pfeiffer2020AdapterHub}\footnote{\url{https://github.com/Adapter-Hub/adapter-transformers}} demonstrates the realization of a large-scale project. This includes tutorials, configurations, and hosting of technical documentation (\url{https://docs.adapterhub.ml/}), as well as a dedicated website for the library itself.

\subsection{Experimental Analysis}\label{app:experimental-analysis}

\paragraph{Statistical Hypothesis Testing} A general introduction to significance testing in NLP is given by \citet{dror2018hitchhiker, raschka2018model, azer2020not}. Furthermore, \citet{dror2020statistical} and \citet{riezler2021validity} provide textbooks around hypothesis testing in an NLP context. \citet{japkowicz2011evaluating} describe the usage of statistical test for general, classical ML classification algorithms. When it comes to usage, \citet{zhang2021cartography} describe the statistical test used with all parameter and results alongside performance metrics.
\citet{shimorina2021error} report p-values alongside test statistics for the Spearman's $\rho$ test, using the Bonferroni correction due to multiple comparisons. \citet{apidianaki2018proceedings} transparently report the p-values of a approximate randomization test \citep{riezler2005some} between all the competitors in an argument reasoning comprehension shared task and interpret them with the appropriate degree of carefulness.

\paragraph{Bayesian analysis} Bayesian Data Analysis has a long history of application across many scientific disciplines. Popular textbooks about the topic are given by \citet{kruschke2010bayesian,gelmanbayesian} with a more gentle introduction by \citet{kruschke2018bayesian}. \citet{benavoli2017time} supply an in-depth tutorial for Bayesian Analysis for Machine Learning, by using a Bayesian signed ranked test \citep{benavoli2014bayesian}, an extension of the frequentist Wilcoxon signed rank test and a Bayesian hierarchical correlated t-test  \citep{corani2015bayesian}. 
Applications can be found for instance by \citet{nilsson2018performance}, who use the Bayesian correlated t-test \citep{corani2015bayesian} to investigate the posterior distribution over the performance difference to compare different federated learning algorithms. To evaluate deep neural networks on road traffic forecasting, \citet{manibardo2021deep} employ Bayesian analysis and plot Monte Carlo samples from the posterior distribution between pairs of models. The plots include ROPEs, i.e., regions of practical equivalence, where the judgement about the superiority of a model is suspended.

\subsection{Publication Considerations}

\paragraph{Replicability}
\citet{gururangan-etal-2020-dont} report in detail all the computational requirements for their adaptation techniques in a dedicated sub-section.
Additionally, following the suggestions by \citet{dodge-etal-2019-show}, the authors report their results on the development set in the appendix.

\paragraph{Environmental Impact}
By introducing MultiBERTs \cite{sellam2021multiberts}, the authors include in their paper an \textit{Environmental Statement}.
In the paragraph they estimate the computational cost of their experiments in terms of hours, and consequential tons of CO2e.
They release the trained models publicly with the aim to allow subsequent studies by other researchers without the computational cost of training MultiBERTs to be incurred.

\citet{hershcovich2022towards}, instead, propose a \emph{climate performance model card} as a way to systematically report the climate impact of NLP research.

\paragraph{Social and Ethical Impact}
\citet{brown-gpt3} present GPT-3 and include a whole section on the \textit{Broader Impacts} language models like GPT-3 have.
Despite improving the quality of text generation, they also have potentially harmful applications.
Specifically, the authors discuss the potential for deliberate misuse of language models, and the potential issues of bias, fairness and representation (focusing on the gender, race and religion dimensions).

The work of \citet{bias_statement} assists the researcher in writing a bias statement, by recommending to provide explicit statements of why the system's behaviors described as ``bias'' are harmful, in what ways, and to whom, then to reason on them. In addition, they provide an example of a bias statement from~\citet{basta2019evaluating}.

\section{Contents of Codebase}\label{app:readme}

\paragraph{The README} First, the initial section of the \texttt{README} would consist of the name of the repository---to what paper or project is this code base tied to? Including a hyperlink to the paper or project itself. Second, developers also indicate the structure of the repository---what and where are the files, folders, code, et cetera in the project and how would they be used. 

Empirical work requires the installation of libraries or software. It is important to install the right versions of the libraries to maintain replicability, and indicate the correct version of the specific package. In Python, a common practice is to make use of virtual environments in combination with a \texttt{requirements.txt} file. The main purpose of a virtual environment is to create an isolated environment for code projects. Each project can have its own dependencies (libraries) regardless of what dependencies every other project has to avoid clashes between libraries. For example, this file can be created by piping the output of \texttt{pip} \texttt{freeze} to a \texttt{requirements.txt} file. For further examples of virtual environment tools, we refer to~\cref{table:resources} (\cref{app:resources}).

To ensure replicability, the practitioner writes a description on how to re-run all experiments that are depicted in a paper to get the same results. For example, these are evaluation scores or graphical plots. This can come in the form of a bash script, that indicates all the commands necessary.\footnote{See for instance \url{https://robvanderg.github.io/blog/repro.htm}} Similarly, one can also indicate all commands in the \texttt{README}.
To give credit to each others work, the last section of the \texttt{README} is usually reserved for credits, acknowledgments, and the citation. The citation is preferably provided in \texttt{BibTeX} format.

\paragraph{Project Structure} 
From the Python programming language perspective, there are several references for initializing an adequate Python project structure.\footnote{Some examples: \url{https://docs.python-guide.org/writing/structure/} and \url{https://coderefinery.github.io/reproducible-research/02-organizing-projects/}} This includes a \texttt{README}, LICENSE, \texttt{setup.py}, \texttt{requirements.txt}, and unit tests. To quote \textit{The Hitchhiker's Guide to Python} \citep{reitz2016hitchhiker} on the meaning of ``structure'':

\begin{quote}
    ``By `structure' we mean the decisions you make concerning how your project best meets its objective. We need to consider how to best leverage Python’s features to create clean, effective code. In practical terms, `structure' means making clean code whose logic and dependencies are clear as well as how the files and folders are organized in the filesystem.''
\end{quote}

This includes decisions on where functions should go into which modules. Also on how data flows through the project. What features and functions can be grouped together or even isolated? In a broader sense, to answer the question on how the finished product should look like.

\section{Resources}\label{app:resources}

An overview over all mentioned resources in the paper is given in \cref{table:resources}.

\begin{table*}
    \tiny 
    \centering 
    \caption{Overview over mentioned resources.} 
    \resizebox{\textwidth}{!}{
            \renewcommand{\arraystretch}{2}%
            \begin{tabularx}{.9\textwidth}{m{2.9cm}XX}
            \toprule
            Name & Description & Link \\
            \midrule
            ACL Anthology  & Website hosting all the published proceedings of the ACL. & \url{https://aclanthology.org}\\%\hline
            ACL pubcheck & Tool to check the format and the citations of papers written with the ACL style files. & \url{https://github.com/acl-org/aclpubcheck}\\
            Anonymous Github  & Website to anonymize a Github repository. & \url{https://anonymous.4open.science}\\%\hline
            baycomp \citep{benavoli2017time} & Implementation of Bayesian tests for the comparison of classifiers. & \url{https://github.com/janezd/baycomp} \\
            BitBucket                  &  A website and cloud-based service that helps developers store and manage their code, as well as track and control changes to their code. & \url{https://bitbucket.org/}\\%\hline
            Conda                    & Open Source package management system and environment management system. & \url{https://docs.conda.io/} \\%\hline
            codecarbon \citep{codecarbon} & Python package estimating and tracking carbon emission of various kind of computer programs. & \url{https://github.com/mlco2/codecarbon} \\%\hline
            dbpl & Computer science bibliography to find correct versions of papers. & \url{https://dblp.org/} \\ %\hline
            deep-significance \citep{ulmer2022deep} & Python package implementing the ASO test by \citet{dror2019deep} and other utilities & \url{https://github.com/Kaleidophon/deep-significance} \\%\hline
            European Language Resources Association~\cite{elra-citation} & Public institution for language and evaluation resources & \url{http://catalogue.elra.info/en-us/}\\
            GitHub                 &  A website and cloud-based service that helps developers store and manage their code, as well as track and control changes to their de. & \url{https://github.com/}\\%\hline
            Google Scholar         & Scientific publication search engine. Note that the ACL Anthology should be preferred, as Google Scholar often indexes the first occurence of a paper (which is frequently a pre-print) & \url{https://scholar.google.com/} \\ %\hline
            Hugging Face Datasets \citep{quentin_lhoest_2021_5510481} & Hub to store and share datasets & {\url{https://huggingface.co/datasets}} \\%\hline
            HyBayes  \citep{azer2020not} & Python package implementing a variety of frequentist and Bayesian significance tests & \url{https://github.com/allenai/HyBayes} \\%\hline
            LINDAT/CLARIN \cite{varadi2008clarin} & Open access to language resources and other data and services for the support of research in digital humanities and social sciences & \url{https://lindat.cz/} \\
            ONNX & Open format built to represent Machine Learning models. & \url{https://onnx.ai/} \\%\hline
            Pipenv                 & Virtual environment for managing Python packages & \url{https://pipenv.pypa.io/}\\%\hline
            Protocal buffers       & Data structure for model predictions & \url{https://developers.google.com/protocol-buffers/} \\%\hline
            \texttt{rebiber} & Python tool to check and normalize the bib entries to the official published versions of the cited papers. & \url{https://github.com/yuchenlin/rebiber} \\ %\hline
            Semantic Scholar & Scientific publication search engine. & \url{https://www.semanticscholar.org/} \\ %\hline
            Virtualenv              & Tool to create isolated Python environments. & \url{https://virtualenv.pypa.io/}\\
            Zenodo                  & General-purpose open-access repository for research papers, datasets, research software, reports, and any other research related digital artifacts & \url{https://zenodo.org/}\\
            
            \bottomrule
        \end{tabularx}%
    }\label{table:resources}
\end{table*}

\end{document}